%% file: main.tex
\documentclass{article}
\usepackage{iclr2027_conference,times}

\input{math_commands.tex}

\usepackage{hyperref}
\hypersetup{pdfborder={0 0 0}}
\usepackage{url}
\usepackage{cancel}
\usepackage{amsmath}
\usepackage{graphicx}
\usepackage{subcaption}
\usepackage{amssymb}
\usepackage{sidecap}
\usepackage{caption}
\usepackage{wrapfig}
\usepackage[ruled,vlined,linesnumbered]{algorithm2e}
\usepackage{float}
\usepackage{placeins}
\usepackage[bottom]{footmisc}
\usepackage{enumitem}
\usepackage{rotating}
\usepackage{booktabs}

\usepackage{amsthm}

\newtheorem{theorem}{Theorem}[section]
\newtheorem{lemma}[theorem]{Lemma}
\newtheorem{proposition}[theorem]{Proposition}

\theoremstyle{definition}

\theoremstyle{remark}

\usepackage[T1]{fontenc}
\usepackage{xcolor}

\title{Distractor-Robust Reinforcement Learning via Variational Bisimulation}

\author{%
Benjamin Freed\thanks{Equal contribution.} \\
Carnegie Mellon University
\And
Vivaan Singh\footnotemark[1] \\
Carnegie Mellon University
\And
Dean Zadok \\
Carnegie Mellon University
\AND
Roberto Calandra \\
Technische Universit\"at Dresden
\And
Jeff Schneider \\
Carnegie Mellon University
\And
Howie Choset \\
Carnegie Mellon University
}

\setlist{topsep=3pt, itemsep=2pt, parsep=1pt}
\usepackage{setspace}
\iclrfinalcopy
\begin{document}
\raggedbottom
\setstretch{0.95}

\maketitle
\lhead{Preprint}

\begin{abstract}
    Model-based reinforcement learning (MBRL) promises data efficiency and generalization, but typical reconstruction-based objectives encourage models to waste representational capacity on task-irrelevant distractors.  We introduce VIBES (Variational Inference for Bisimulation-based Encoded States), a new objective that replaces pixel reconstruction with an adversarial term, which enforces that latent states suffice to predict both rewards and the \emph{next} latent state. We show that this objective measures how far the encoder is from reward and dynamics sufficiency, and that whenever exact sufficiency is attainable, its global optima induce bisimulation relations on histories. Combined with a fixed-capacity latent bottleneck, this encourages latent states to retain task-relevant information while discarding irrelevant variation. Our method serves as a drop-in replacement for Dreamer’s model-learning component, achieves state-of-the-art performance on the Distracting Control Suite, and extends to visually complex manipulation in Realistic ManiSkill. Unlike prior approaches, it does not rely on image-specific augmentations and applies equally well to high-dimensional vector-state tasks, demonstrated on a 100-link swimmer robot. Finally, latent-space analyses (UMAP embeddings and nearest-neighbor probes) confirm that the learned representations are sensitive to task-relevant structure while invariant to distractors.

\end{abstract}

\section{Introduction}
\label{sec:intro}
Model-based reinforcement learning (MBRL) promises a powerful set of advantages: it can achieve high data efficiency by reusing learned dynamics for many policy updates, support transfer of knowledge across tasks through shared models, enable long-horizon reasoning and counterfactual planning, and, when paired with uncertainty estimates, drive intelligent exploration. These benefits make MBRL an attractive alternative to model-free methods.

However, in practice, learned models are inevitably imperfect—limited by capacity and the cost of collecting data and performing optimization—and this discrepancy between the model and the true environment often leads to suboptimal policies.  One might hope that at least model-based RL methods would consistently find the "best" of the available set of imperfect candidate models; however, this is not even the case, because typical model training objectives are not aligned with the ultimate goal of improving control performance.  This issue is referred to as objective mismatch \citep{objective_mismatch}. Worse still, typical objectives actively incentivize models to expend representational capacity on irrelevant \textit{distractors}—features of the observation that have no bearing on the task.

Consider the case of autonomous driving: while the agent must model the road, traffic signals, and surrounding vehicles, it need not track details such as the positions of leaves on nearby trees. Yet typical model-learning objectives reward the model for reconstructing those leaves just as faithfully as traffic lights, thereby wasting capacity on irrelevant information. This problem has been shown to significantly degrade the performance of otherwise strong algorithms such as Dreamer \citep{dreamer}, which performs well on clean image-based benchmarks but struggles in the presence of distractors \citep{tpc,repo}.  This phenomenon exemplifies objective mismatch because the agent actively chooses a model that is poor for policy improvement, even if a model that ignores distractors and is well-suited to policy improvement exists within its model class. 

A natural way to mitigate objective mismatch is through state abstraction. State abstraction filters observations so that only task-relevant information is retained. In MBRL, this allows a dynamics model to focus its capacity on predicting what matters for decision-making, rather than reconstructing irrelevant features. Numerous approaches to state abstraction have been proposed, but many rely on heuristic, domain-specific strategies, such as image augmentations \citep{curl,drq,drqv2,rad,netrand}. This partially defeats the purpose of RL, which is to solve tasks generally, without relying on domain knowledge.

In this work, we instead pursue a principled approach based on \emph{bisimulation relations} \citep{algebraic_laws,givan2003equivalence}. A bisimulation relation groups states that yield identical immediate rewards and identical distributions over future groups; the coarsest such relation retains exactly the task-relevant information while discarding irrelevant variation. Our work builds on recent approaches that use \emph{bisimulation metrics}—continuous distance functions that approximate behavioral equivalence \citep{zhanglearning,castro2020scalable}. In contrast, we target bisimulation \emph{relations} directly. We show that our variational objective decomposes into a measure of how far the encoder is from reward and dynamics sufficiency, and that whenever exact sufficiency is attainable, its global optima induce bisimulation relations on histories. A fixed-capacity latent bottleneck then pressures the encoder toward coarse relations that discard distractors. This yields a representation aligned with the theoretical definition of bisimulation while avoiding the need for brittle and hyperparameter-sensitive distance metrics.  We call our resulting model-based RL algorithm \textbf{VIBES}: \emph{Variational Inference for Bisimulation-based Encoded States}.

We evaluate our method on the challenging Distracting Control Suite (DCS) \citep{dcs}, where standard DeepMind Control Suite tasks are augmented with strong visual distractors, and on Realistic ManiSkill \citep{repo}, where robotic manipulation tasks from ManiSkill2 \citep{maniskill2} are set against real indoor scenes from Matterport3D \citep{matterport3d}. On DCS, our approach achieves the highest mean return across tasks, significantly outperforming prior state-of-the-art RL methods, both model-free and model-based, for distractor robustness (Welch's t-test, $p < 0.05$), despite requiring no specialized data augmentation. On Realistic ManiSkill, it shows no significant difference from DreamerV3-12M \citep{dreamerv3} while using fewer than one-third as many learned parameters. Moreover, we demonstrate that the learned representations generalize to unseen distractors and also yield strong performance in a high-dimensional vector-state control task (a swimmer robot with a 401-dimensional observation space). Finally, latent-space analyses confirm that our method learns representations that preserve task-relevant structure, discard distractors, and capture symmetries in the dynamics (e.g., rotational invariances), yielding a more compact and general abstraction.

In summary, the contributions of this paper are as follows:
\begin{enumerate}
    \item We propose a principled state abstraction method based on bisimulation relations, integrated into a variational inference framework for model learning. We show that its objective measures how far the encoder is from reward and dynamics sufficiency, and that its global optima induce bisimulation relations whenever exact sufficiency is attainable.
 \item We demonstrate strong empirical performance on image-based tasks with heavy distractors (DCS and Realistic ManiSkill) and on high-dimensional vector-state control, without requiring data augmentations.
    \item We provide quantitative analyses showing that our learned representations preserve task-relevant features while discarding irrelevant variation.
\end{enumerate}

\begin{figure}[t]
  \centering
  \subcaptionbox{PGM assumed by \textit{Dreamer}.
    \label{fig:pgm-dreamer}}%
    [0.40\linewidth]{\includegraphics[width=\linewidth]{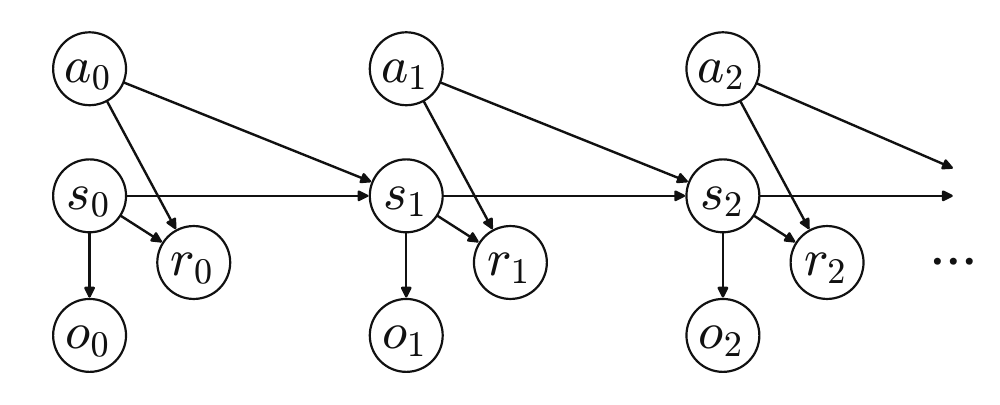}}
  \hfill
  \subcaptionbox{PGM assumed by \textit{VIBES}.
    \label{fig:pgm-ours}}%
    [0.40\linewidth]{\includegraphics[width=\linewidth]{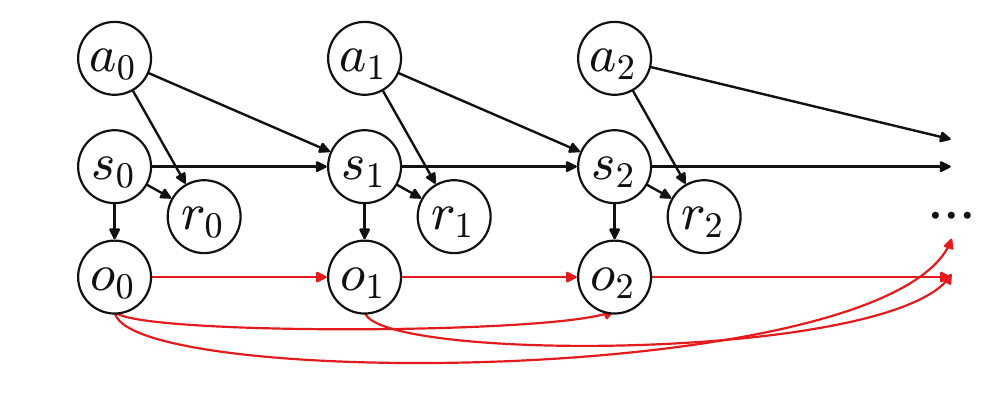}}
  \caption{Side-by-side comparison of graphical model assumptions.
  These Probabilistic Graphical Models (PGMs) model environment dynamics, where the latent state $\tilde{s}_t$ represents the true, underlying hidden state from which observations and rewards arise. Actions may be either fixed exogenously or depend on observations through a policy.
  \textbf{Left (Dreamer):} the observation at time $t$ is conditionally independent of prior observations and actions given the current latent state, i.e., $o_t \perp \!\!\! \perp o_{<t},a_{<t}\mid \tilde{s}_t$.
  \textbf{Right (Ours):} we allow $o_t$ to depend on the full history of observations (and, in general, on the history of states/actions).
  Under Dreamer’s assumption, any information in $o_{<t}$ needed to predict $o_t$ must be encoded into $s_t$; in our formulation this pressure is removed from $s_t$ because past observations can inform $o_t$ directly.}
  \label{fig:pgm-side-by-side}
\end{figure}

\section{Background}

\subsection{Variational Inference for Learning State Representations}

Variational inference (VI) provides a general framework for learning latent-variable models. In such models, observations are assumed to be generated from unobserved latent variables through a probabilistic process. Latent-variable models are powerful because they enable compact, structured representations that capture essential features of the data.  In the context of RL, VI enables learning latent state representations together with dynamics and reward models. Dreamer\footnote{Unless a version is specified, ``Dreamer'' refers to the variational formulation that the Dreamer family builds on \citep{planet,dreamer,dreamerv2,dreamerv3}. Empirically, we compare against the strongest applicable Dreamer variant (Sec.~\ref{sec:dcs}).} is a prominent VI formulation that maximizes a lower bound on $\log p(o_{0:T}, r_{0:T}\mid a_{0:T-1})$ using a decoder $\hat p(o_t\mid s_t)$, a transition $\hat p(s_t\mid s_{t-1},a_{t-1})$, and a reward model $\hat p(r_t\mid s_t,a_t)$. A key modeling assumption is $o_t \perp\!\!\!\perp (o_{<t},a_{<t}) \mid s_t$, which pressures $s_t$ to carry any history information useful for pixel prediction—even if irrelevant for control (Figure \ref{fig:pgm-dreamer}). This “remember everything” incentive misaligns model learning with decision making, especially when in tasks with distractors.

In Sec. \ref{sec:learning_bisim_relations}, we propose a modified objective that relaxes this conditional independence assumption. By allowing observations to depend on past observations directly, we remove the pressure for $\tilde{s}_t$ to transmit irrelevant predictive information, freeing the latent representation to focus on features that matter for predicting rewards and dynamics (Fig. \ref{fig:pgm-ours}).

\subsection{Bisimulation Relations}

Bisimulation originates in the theory of Markov decision processes (MDPs) and provides a principled notion of when two states can be considered behaviorally equivalent. Intuitively, two states are bisimilar if they yield the same rewards and, under any action, transition to future states that are themselves bisimilar. This recursive definition ensures that bisimilar states are indistinguishable for the purposes of planning and control.

In our setting, the agent does not observe the environment state directly, only observations from which the state must be inferred. We therefore work in a partially observable MDP (POMDP) with action space $\mathcal{A}$. In a POMDP, the history $H_t = (o_{\le t}, a_{<t})$ plays the role of the state~\citep{ni2024bridging, simao2023safe}, so a bisimulation relation partitions the space of all histories  into groups of equivalent histories.

Formally, let $\mathcal{H}$ denote the space of all histories. An equivalence relation $\sim$ on $\mathcal{H}$ is a \emph{bisimulation relation} if, for all $H^i, H^j \in \mathcal{H}$,
\begin{equation}
\label{eq:bisim-def}
H^i \sim H^j \;\implies
\begin{cases}
R(H^i, a) = R(H^j, a) & \forall a \in \mathcal{A},\\
P_{\tilde{H}' \mid H, a}(\tilde{H}' \mid H^i, a) = P_{\tilde{H}' \mid H, a}(\tilde{H}' \mid H^j, a) & \forall a \in \mathcal{A},\ \forall \tilde{H}' \in \tilde{\mathcal{H}},
\end{cases}
\end{equation}
where $\tilde{\mathcal{H}}$ is the set of all groups of equivalent histories and $\tilde{H}'$ is a particular group of equivalent histories; $P_{\tilde{H}' \mid H, a}(\tilde{H}' \mid H, a) = \sum_{H' \in \tilde{H}'} P(H' \mid H, a)$ is the probability that the next history falls in group $\tilde{H}'$, where $P(H' \mid H, a)$ is the probability of observing $o'$ after taking action $a$ given history $H$, so that the next history is $H' = (H, a, o')$; and $R(H, a)$ is the distribution over the reward received after taking action $a$ given history $H$. Put differently, equivalent histories must yield identical immediate reward distributions under every action and identical distributions over groups of next histories.

 Bisimulation relations are not unique: the identity relation, which distinguishes every pair of histories, satisfies \eqref{eq:bisim-def} trivially. The largest bisimulation relation induces the coarsest partition of $\mathcal{H}$, whose equivalence classes can be treated as abstract states without loss of optimal control performance~\citep{givan2003equivalence, li2006towards}, making bisimulation an ideal basis for a state representation.

\paragraph{Bisimulation for RL} Bisimulation is appealing for reinforcement learning because it formalizes the idea of \emph{state abstraction}: discarding irrelevant details of the environment while preserving task-relevant information. A representation that encodes the equivalence classes of the coarsest bisimulation retains precisely the information needed for reward prediction and decision making. By contrast, states that differ only in irrelevant features (e.g., background distractors) are collapsed into the same abstract state.

Recent work has sought to approximate bisimulation in continuous or high-dimensional settings by learning embeddings that preserve bisimulation distances \citep{ferns2004metrics, zhanglearning}. We refer the reader to Appendix~\ref{app:relatedwork} for an extended review of the landscape of comparable representation-learning methods. In practice, exact bisimulation is infeasible in complex environments, but the principle provides a powerful guide: representations should be invariant to irrelevant variation while distinguishing states that differ in reward or in their distribution over future behaviors. In this work, we show how to incorporate this principle directly into a variational inference framework for model-based RL, yielding latent representations that approximate bisimulation relations and are robust to distractors.

\section{Learning Bisimulation Relations with Variational Inference}
\label{sec:learning_bisim_relations}

\begin{figure}[t]
  \centering
  \includegraphics[width=0.88\linewidth]{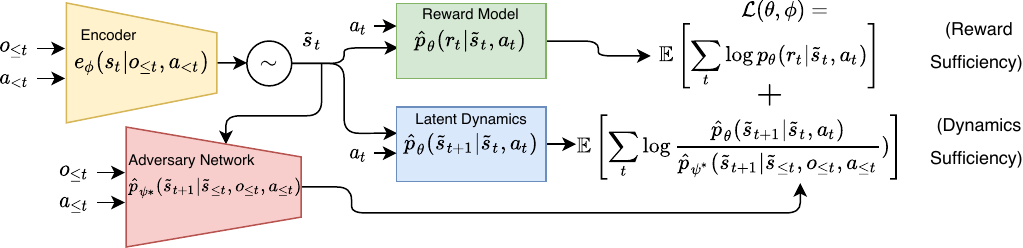}
  \caption{\textbf{Model diagram and objective.}
  Our model comprises an \emph{encoder} (CNN+GRU) that maps history $(o_{\le t},a_{<t})$ to a stochastic latent $\tilde{s}_t$; a \emph{dynamics model} that predicts $\tilde{s}_{t+1}$ from $(\tilde{s}_{t},a_{t})$; a \emph{reward model} $p(r_t \mid \tilde{s}_t, a_t)$; and an \emph{adversary network} that predicts $\tilde{s}_{t+1}$ from the history and current action, $(o_{\le t},a_{\le t})$.
  Our objective has two parts: a \textit{reward sufficiency} term, maximized when the latent state contains all information needed to predict the current reward, and a \textit{dynamics sufficiency} term, maximized when the latent state is as informative as the full history for predicting the \emph{next} latent state.}
  \label{fig:model-diagram}
\end{figure}

We adopt a variational approach to learn a latent state, a dynamics model, and a reward model, but replace pixel reconstruction with an \emph{adversarial} term that measures whether the latent state at time $t$ is as informative as the full history for predicting the \emph{next} latent state. The resulting objective has two parts: a \emph{reward sufficiency} term (the latent must predict rewards) and a \emph{dynamics sufficiency} term (the latent must be as predictive of the next latent as history). A full derivation is deferred to Appendix~\ref{app:model_obj_derivation}.

\subsection{Model Objective}
\label{sec:model-objective}
Let $\tilde{s}_t \sim e_{\phi}(o_{\le t}, a_{<t})$ denote the stochastic latent at time $t$, where \(e_{\phi}\) is an \textit{encoder network}. Our encoder outputs a Gaussian with fixed standard deviation $\sigma_e = 0.01$ and a unit-norm mean, making the latent a finite-capacity channel. The objective (Section~\ref{sec:theory}) rewards retaining the information needed to predict rewards and future latents but does not penalize retaining more; this bottleneck supplies the pressure to discard distractors.

Our model objective is
\begin{equation}
\label{eq:model_loss}
\mathcal{L}(\theta,\phi)\;=\;
\mathbb{E}\!\left[
\sum_{t}
\log\frac{\hat{p}_{\theta}(\tilde{s}_{t}\mid \tilde{s}_{t-1},a_{t-1})}
{\hat{p}_{\psi^*}(\tilde{s}_{t}\mid \tilde{s}_{<t}, o_{<t}, a_{<t})}
\;+\;
\log \hat{p}_{\theta}(r_t\mid \tilde{s}_t,a_t)
\right].
\end{equation}
The full derivation is given in Appendix~\ref{app:model_obj_derivation}. Here $\hat{p}_{\theta}(\tilde{s}_{t}\mid \tilde{s}_{t-1},a_{t-1})$ is the latent dynamics;
$\hat{p}_{\theta}(r_t\mid \tilde{s}_t,a_t)$ is the reward model; and
$\hat{p}_{\psi^*}(\tilde{s}_{t}\mid \tilde{s}_{<t}, o_{<t}, a_{<t})$ is an \emph{adversary} (the optimal history-conditioned predictor of $\tilde{s}_t$), obtained by an inner maximization. In practice, we approximate $\psi^*$ by training the adversary with a higher learning rate than $\phi$, so that $\psi$ tracks close to its optimum at each step. Our analysis treats $\tilde{s}_t$ as a Markov latent state; in our implementation, the dynamics model, actor, and critic use an augmented state $(h_t,\tilde{s}_t)$, with $h_t$ a deterministic recurrent summary of past latents and actions . Our theoretical results hold unchanged for this augmented state, as shown in App.~\ref{app:augmented}.

\paragraph{Intuitive explanation of \(\mathcal{L}\).}  The reward sufficiency term \(\log p(r_t|\tilde{s}_t,a_t) \) encourages $\tilde{s}_t$ to be sufficient for predicting $r_t$ given $a_t$.
The dynamics sufficiency term \( \frac{\hat{p}_{\theta}(\tilde{s}_{t}|\tilde{s}_{t-1},a_{t-1})}{\hat{p}_{\psi*}(\tilde{s}_{t}| \tilde{s}_{< t}, o_{< t},a_{< t})} \) compares two predictors of $\tilde{s}_t$: our latent dynamics model \(\hat{p}_{\theta}(\tilde{s}_{t}|\tilde{s}_{t-1},a_{t-1})\) that only sees $(\tilde{s}_{t-1}, a_{t-1})$ vs.\ an optimal predictor that sees the full history, \(\hat{p}_{\psi*}(\tilde{s}_{t}| \tilde{s}_{< t}, o_{< t},a_{< t})\). Even though $\tilde{s}_{t-1}$ is itself built from the full history by the encoder, it is a (potentially) lossy compression of that history, whereas the adversary observes the raw history directly. This makes the adversary a useful training-time signal: whenever the encoder has discarded task-relevant information, the adversary can still recover it and out-predict the dynamics model, and this gap is exactly what pressures the encoder to retain that information. With an optimal adversary, its expectation is at most $0$, with equality exactly when $\tilde{s}_{t-1}$ contains all history information needed to predict $\tilde{s}_t$ and the dynamics model is correct (Theorem~\ref{thm:approx}).
Maximizing \eqref{eq:model_loss} therefore pushes the representation toward \emph{reward} and \emph{dynamics} sufficiency.

\subsection{Why \(\mathcal{L}\) yields bisimulation}
\label{sec:theory}

Full proofs are in App.~\ref{app:proofs}. Throughout, $p(\cdot)$ denotes true distributions, obtained by sampling trajectories from the training data and passing them through the encoder, and $\hat{p}(\cdot)$ denotes our learned models (dynamics, reward, and adversary). We assume the adversary is optimal, so $\hat{p}_{\psi^*}(\tilde{s}_t\mid\tilde{s}_{<t},o_{<t},a_{<t}) = p(\tilde{s}_t\mid\tilde{s}_{<t},o_{<t},a_{<t})$ (\eqref{eq:psi_objective}).  An encoder groups histories: $H \sim_{e_\phi} H'$ iff $e_\phi(\cdot\mid H) = e_\phi(\cdot\mid H')$, i.e., the encoder maps both histories to the same latent distribution. For our fixed-variance Gaussian encoder, this groups histories by the encoder mean. We ask when this grouping is a bisimulation.

\begin{lemma}[Sufficiency induces bisimulation]
\label{lem:sufficiency}
Suppose there exist kernels $\kappa$ and $\rho$ such that, for every history $H$, action $a$, and value $s$ in the support of $e_\phi(\cdot\mid H)$, the next latent $\tilde{s}'$ and the reward $r$ satisfy
\begin{align}
\textbf{(Dynamics sufficiency)}\quad & P(\tilde{s}' \mid H, \tilde{s} = s, a) = \kappa(\tilde{s}' \mid s, a), \label{eq:dyn-suff}\\
\textbf{(Reward sufficiency)}\quad & P(r \mid H, \tilde{s} = s, a) = \rho(r \mid s, a), \label{eq:rew-suff}
\end{align}
and that the encoder family is identifiable (App.~\ref{app:lemma_proof}; this holds for deterministic encoders and for Gaussian encoders with fixed covariance). Then $\sim_{e_\phi}$ is a bisimulation relation. (Given the current latent, the history carries no further information about the next latent or the reward.)
\end{lemma}

\begin{theorem}[Global optima yield bisimulation under realizability]
\label{thm:exact}
Suppose the encoder and model classes are \emph{jointly realizable}: some encoder $\phi^\circ$ and model $\theta^\circ$ make both predictions exact, i.e., for all $t$,
\begin{equation}
\hat{p}_{\theta^\circ}(\tilde{s}_t\mid\tilde{s}_{t-1},a_{t-1}) = p(\tilde{s}_t\mid\tilde{s}_{<t},o_{<t},a_{<t}),
\qquad
\hat{p}_{\theta^\circ}(r_t\mid\tilde{s}_t,a_t) = p(r_t\mid o_{\le t},a_{\le t}).
\label{eq:realizable}
\end{equation}\footnote{These equalities hold $p$-almost surely, i.e., for all histories, actions, and latents that occur with nonzero probability under $p$; App.~\ref{app:theorem_proof} gives the precise statement.} Then every global maximizer $(\theta^*,\phi^*)$ of \eqref{eq:model_loss} also satisfies \eqref{eq:realizable}. If, moreover, the training data covers all histories and actions and the encoder family is identifiable, then $e_{\phi^*}$ satisfies the sufficiency conditions (equations~\ref{eq:dyn-suff}--\ref{eq:rew-suff}), and by Lemma~\ref{lem:sufficiency}, $\sim_{e_{\phi^*}}$ is a bisimulation relation.
\end{theorem}

Theorem~\ref{thm:exact} requires exact sufficiency to be attainable. More generally, define the \emph{sufficiency violation} of an encoder,
\begin{equation}
\Delta(\phi) := \sum_t \Big[ I\big(\tilde{s}_t;\, \tilde{s}_{<t},o_{<t},a_{<t} \,\big|\, \tilde{s}_{t-1},a_{t-1}\big) + I\big(r_t;\, o_{\le t},a_{\le t} \,\big|\, \tilde{s}_t,a_t\big)\Big] \;\ge\; 0,
\label{eq:delta}
\end{equation}
the information the history carries about the next latent and the reward beyond what the latent carries; it is zero exactly when both sufficiency conditions hold.

\begin{theorem}[$\mathcal{L}$ measures sufficiency violation]
\label{thm:approx}
 Assume that, given the history, rewards have a density with finite entropy.\footnote{Deterministic rewards can be accommodated by adding arbitrarily small noise to the reward.} Then for all $\theta,\phi$, $\mathcal{L}(\theta,\phi) = C - \Delta(\phi) - M(\theta,\phi)$, where $C = \sum_t \mathbb{E}[\log p(r_t\mid o_{\le t},a_{\le t})]$ is finite and does not depend on $(\theta,\phi)$, and $M(\theta,\phi) \ge 0$ is the expected KL divergence from the best latent-conditioned predictors, $p(\tilde{s}_t\mid\tilde{s}_{t-1},a_{t-1})$ and $p(r_t\mid\tilde{s}_t,a_t)$, to the dynamics and reward models. Hence $\mathcal{L}(\theta,\phi) \le C - \Delta(\phi)$, with equality when the model class contains these predictors; in that case, maximizing $\mathcal{L}$ is equivalent to minimizing $\Delta$ over the encoder class.
\end{theorem}

 For encoders with full-support noise such as ours, exact sufficiency is unattainable (Proposition~\ref{prop:fixed-sigma}). Theorem~\ref{thm:exact} thus describes an idealized limit, while Theorem~\ref{thm:approx} governs the implemented method: maximizing $\mathcal{L}$ drives the encoder toward minimal sufficiency violation, up to the expressiveness of the dynamics and reward models. Neither result implies that distractors are discarded, since an encoder that preserves the full history is also sufficient; that pressure comes from the bottleneck (Section~\ref{sec:model-objective}), and we evaluate it empirically (Section~\ref{sec:latent-analysis}).

\paragraph{Optimizing the Model Objective} Optimizing \eqref{eq:model_loss} is more involved than in Dreamer because $\psi^*$ is defined by an inner maximization (\eqref{eq:psi_objective}) and therefore depends on the model parameters. Differentiating through this dependence would require solving a bilevel problem. However, by the envelope theorem, this dependence contributes nothing to the gradient (Proposition~\ref{prop:envelope}, App.~\ref{app:envelope}): since $\psi^*$ already maximizes the adversary's objective, small changes in $\psi^*$ leave that objective unchanged to first order. Hence $\nabla_{\theta}\mathcal{L}$ and $\nabla_{\phi}\mathcal{L}$ equal the gradients of \eqref{eq:model_loss} with $\psi$ held fixed at $\psi^*$. Only the adversary's \emph{parameters} are held fixed; the encoder still receives gradients through the latents $\tilde{s}$ that the adversary takes as inputs and predicts as targets. This reduces model training to an alternating optimization, $\theta$ and $\phi$ take a gradient step with $\psi$ frozen, and $\psi$ is then updated toward its optimum. The resulting gradient is exact when $\psi = \psi^*$, so we train the adversary with a higher learning rate than $\phi$ so that $\psi$ tracks its optimum; in practice, one adversary step per model step was sufficient in preliminary experiments; Appendix~\ref{app:adversary-gap} reports the adversary--dynamics likelihood gap over training. Our complete model training procedure is summarized in Algorithm~\ref{alg:model_training} in Appendix~\ref{app:model_training}.

\paragraph{Integration into Online Reinforcement Learning}

The objective in \eqref{eq:model_loss} serves as a direct drop-in replacement for the model-learning component of Dreamer. 
Once trained, the resulting world model is used in a manner similar to Dreamer: to simulate imagined rollouts in the latent state space for policy and value learning.
Thus, apart from replacing the reconstruction loss with our bisimulation-driven objective, the rest of the Dreamer framework (policy optimization, value estimation, and training loop) remains largely unchanged. Specific details of the RL training framework are presented in Appendix~\ref{app:model_arch}.

\section{Experiments}

We evaluate VIBES in four settings. We first benchmark it on the Distracting Control Suite (DCS) and ablate the adversarial objective (Section~\ref{sec:dcs}). We then test robotic manipulation under realistic background distractors on Realistic ManiSkill (Section~\ref{sec:realistic-maniskill}) and high-dimensional control from vector states (Section~\ref{sec:vector-states}). Finally, we analyze the learned latent spaces to test whether they preserve task-relevant features while discarding distractors (Section~\ref{sec:latent-analysis}). Architectures, hyperparameters, and benchmark configurations are in Appendix~\ref{app:implementation}.

\subsection{Distracting Control Suite}
\label{sec:dcs}
\begin{table}[t]
\centering
{\fontsize{7.6}{8.4}\selectfont
\renewcommand{\arraystretch}{0.9}
\begin{tabular}{lccccccc}
\hline
Method              & Mean         & BiC-Catch    & C-swingup    & C-run        & F-spin       & R-easy       & W-walk \\
\hline
SAC+RAD$\dagger$    & 270 $\pm$ 55 & 366 $\pm$ 116 & 297 $\pm$ 41 & 198 $\pm$ 76 & 338 $\pm$ 116  & 173 $\pm$ 22 & 249 $\pm$ 271 \\
QT-Opt+RAD$\dagger$ & 343 $\pm$ 43 & 490 $\pm$ 125 & 467 $\pm$ 24 & 170 $\pm$ 16  & 393 $\pm$ 178  & 428 $\pm$ 133 & 109 $\pm$ 24 \\
QT-Opt+DrQ$\dagger$ & 265 $\pm$ 25  & 395 $\pm$ 76 & 431 $\pm$ 35 & 126 $\pm$ 20 & 203 $\pm$ 65  & 343 $\pm$ 104 & 91 $\pm$ 6 \\
RSAC$\dagger$       & 275 $\pm$ 25 & 181 $\pm$ 63 & 465 $\pm$ 41 & 292 $\pm$ 20 & 86 $\pm$ 108   & 145 $\pm$ 61 & 482 $\pm$ 39 \\
CURL$\dagger$       & 391 $\pm$ 22 & 102 $\pm$ 39 & 432 $\pm$ 29 & 233 $\pm$ 26 & \underline{648 $\pm$ 63}  & 253 $\pm$ 78 & \underline{678 $\pm$ 69} \\
CoRe$\dagger$       & 586 $\pm$ 38 & \underline{798 $\pm$ 59} & 499 $\pm$ 43 & 423 $\pm$ 43 & \textbf{713 $\pm$ 159} & 340 $\pm$ 118 & \textbf{744 $\pm$ 78} \\
SAR                 & 173 $\pm$ 17 & 286 $\pm$ 84 & 351 $\pm$ 29 & 240 $\pm$ 70 & 14  $\pm$ 16 & 108  $\pm$ 31 & 36 $\pm$ 3 \\
DBC & 68 $\pm$ 12 & 99 $\pm$ 45 & 119 $\pm$ 21 & 62 $\pm$ 49 & 10 $\pm$ 7 & 89 $\pm$ 21 & 29 $\pm$ 2 \\
HRSSM & 599 $\pm$ 58 & \underline{591 $\pm$ 247} & \underline{651 $\pm$ 13} & \underline{454 $\pm$ 34} & \underline{688 $\pm$ 218} & 598 $\pm$ 63 & 611 $\pm$ 82 \\
DreamerV3-12M & 277 $\pm$ 10 & 103 $\pm$ 36 & 339 $\pm$ 10 & 309 $\pm$ 42 & 33 $\pm$ 15 & 332 $\pm$ 17 & 547 $\pm$ 7 \\
VIBES (Ours)        & \textbf{701 $\pm$ 29} & \textbf{844 $\pm$ 95} & \textbf{696 $\pm$ 50} & \textbf{489 $\pm$ 22} & \underline{670 $\pm$ 167} & \textbf{930 $\pm$ 12} & 574 $\pm$ 56 \\[1.5pt]
\hline
No Adv. Obj.        & 606 $\pm$ 27 & 593 $\pm$ 158 & \underline{658 $\pm$ 28} & 436 $\pm$ 27 & 469 $\pm$ 57 & \underline{925 $\pm$ 21} & 556 $\pm$ 23 \\[1.5pt]
Recon. Obj.         & 417 $\pm$ 49 & 144 $\pm$ 96 & 543 $\pm$ 72 & \underline{469 $\pm$ 11} & 347 $\pm$ 164 & 360 $\pm$ 280 & \underline{639 $\pm$ 101} \\[1.5pt]
\hline
\end{tabular}
}
\caption{Performance on Distracting Control Suite tasks at 500K environment steps. Mean episode return $\pm$ 95\% confidence interval across 5 seeds. Bold marks the highest mean in each column; underlined results are not significantly different from the bolded entry (Welch's t-test, $p > 0.05$; full $p$-values in Appendix~\ref{app:significance-tests}). $\dagger$ denotes results taken from prior work, converted from mean $\pm$ SE to 95\% confidence intervals: RSAC, CURL and CoRe from \citet{core} (10 seeds); SAC+RAD, QT-Opt+RAD and QT-Opt+DrQ from \citet{dcs} (5 seeds), as reported in \citet{core}.}
\label{tab:results}
\end{table}

The Distracting Control Suite (DCS) \citep{dcs} extends the DeepMind Control Suite with three types of distractors: (i) \emph{background distractors}, where high-resolution natural videos play behind the agent, (ii) \emph{color distractors}, where the agent’s body color changes dynamically, and (iii) \emph{camera distractors}, where the camera viewpoint shifts during the episode. Each task requires fine-grained control of the agent’s state while the distractors add large amounts of task-irrelevant variation, so DCS directly tests whether an algorithm can overcome the objective mismatch problem from Section~\ref{sec:intro}. Our DCS configuration is detailed in Appendix~\ref{app:dcs_config}; learning curves are shown in Appendix~\ref{app:dcs-learning-curves}.

We compare against SAC+RAD \citep{rad}, QT-Opt \citep{qtopt} variants with RAD and DrQ \citep{drq} regularization, RSAC \citep{rsac}, CURL \citep{curl}, CoRe \citep{core}, SAR \citep{sar}, the bisimulation-based methods DBC \citep{zhanglearning} and HRSSM \citep{hrssm}, and DreamerV3 \citep{dreamerv3}, the strongest Dreamer variant applicable to our setting. We exclude DreamerV4 \citep{dreamerv4}, which targets a different problem: learning multi-task agents from offline human demonstrations. All methods are evaluated at 500K environment steps on a held-out set of background video distractors, directly testing the generalization of the learned representation.

As shown in Table~\ref{tab:results}, VIBES achieves the highest mean return across tasks and the highest per-task mean on four of six tasks. Together with its no-adversary ablation, it is the only method to solve \emph{Reacher-easy}, even though this task is structurally simpler than the others. We hypothesize that this stems from Reacher-easy’s randomized goal location. Many existing methods learn representations that emphasize agent-centric dynamics—the aspects of the system directly influenced by actions. Since the goal position is unaffected by actions, it is often underrepresented in the latent even though it determines the reward. VIBES instead requires the latent to preserve all reward-predictive information, so both the agent state and the goal location are encoded.

\paragraph{Ablation Studies}
Removing the adversarial objective (No Adv.\ Obj.) lowers the mean return on every task, confirming that the term provides a measurable benefit, although the ablation still has a higher mean return than every baseline. We hypothesize that the adversarial term matters least when the task-relevant state is (1) governed by deterministic dynamics and (2) fully observable from the observations. Under these conditions, the encoder mean $\mu_\phi(o_{\le t},a_{<t})$ is a deterministic function of the past, so $p(\tilde{s}_t\mid\tilde{s}_{<t},o_{<t},a_{<t}) = \mathcal{N}(\mu_\phi(o_{\le t},a_{<t}),\sigma_e^2 I)$ has a fixed entropy. The adversarial objective therefore reduces to a constant and can be omitted without changing the optimal solution. Condition (1) holds in our setting because the underlying MuJoCo simulator is deterministic. Condition (2) can fail when limited rendering resolution or distractors obscure the agent's state (e.g., when background or camera motion overlaps with agent movement). The adversarial term is therefore most valuable where observations are imperfectly informative about the underlying state, which explains its empirical benefits.

We also evaluate Recon.\ Obj., which keeps our architecture but replaces the adversarial objective with a Dreamer-style pixel reconstruction objective: \\
\(
\mathcal{L}_{\textrm{recon}}(\theta,\phi) 
= \mathbb{E}_{\vec{o},\vec{r},\vec{a},\vec{\tilde{s}}}  \left[ 
\sum_t \left( \log \hat{p}_{\theta}(o_t|\tilde{s}_t) + \log \hat{p}_{\theta}(r_t|\tilde{s}_t,a_t)
+ \log \frac{\hat{p}_{\theta}(\tilde{s}_t|\tilde{s}_{t-1},a_{t-1})}{ \textcolor{gray}{e_{\phi}(\tilde{s}_t|o_{\le t},a_{<t})} } \right) \right] .
\)
\footnote{Because our encoder uses a fixed variance, the term \textcolor{gray}{in gray} is constant and therefore omitted during optimization.\label{fn:gray-term}}

Since this is the closest analogue of Dreamer’s objective that our architecture admits, comparing it with VIBES isolates the contribution of the objective. Replacing the adversarial objective with reconstruction substantially lowers mean performance, though \emph{Cheetah-run} is nearly unaffected and \emph{Walker-walk} improves; when reconstruction remains competitive is an open question. Recon. Obj. is not a Dreamer model: it inherits our architecture\footnote{Our encoder outputs a unit-norm mean with fixed standard deviation and uses a dedicated GRU. In DreamerV1, the posterior mean and standard deviation are learned without a norm constraint, while DreamerV2 and V3 use categorical latents. In all three, the posterior shares the dynamics model’s recurrent state. Full architectural details are in Appendix~\ref{app:model_arch}.} and, because the encoder's standard deviation is fixed, drops the encoder entropy term from the KL (footnote~\footref{fn:gray-term}). These differences may explain why it outperforms DreamerV3~\citep{dreamerv3}.

\FloatBarrier
\subsection{Realistic ManiSkill}
\label{sec:realistic-maniskill}

Following HRSSM \citep{hrssm}, we evaluate VIBES on two tasks from Realistic ManiSkill \citep{repo}, a variant of ManiSkill2 \citep{maniskill2}: Lift Cube and Turn Faucet, which require controlling a robot arm to lift a cube above a target height and rotate a faucet handle to a target angle, respectively. The default plain scene behind the robot and object is replaced with real indoor scenes from the Matterport3D dataset \citep{matterport3d}, sampled from a pool of 88 scenes at each episode reset and held fixed within the episode. This tests whether VIBES uses model capacity for task-relevant information rather than background reconstruction. Full details of this task are provided in Appendix~\ref{app:maniskill-protocol}.

\begin{table}[H]
\centering
{\fontsize{7.6}{8.4}\selectfont
\renewcommand{\arraystretch}{0.9}
\begin{tabular}{lrrr}
\hline
Method & Learned params. & Lift Cube & Turn Faucet \\
\hline
TIA$\dagger$               & 13.750M      & $\underline{274 \pm 196}$ & $47 \pm 26$ \\
Denoised MDP$\dagger$      & 7.483M   & $155 \pm 80$  & $71 \pm 31$ \\
RePo$\dagger$              & 1.671M            & $83 \pm 18$   & $92 \pm 70$ \\
HRSSM$\dagger$             & 10.991M           & $284 \pm 68$  & $278 \pm 78$ \\
DreamerV3-1M               & 0.687M            & $434.50 \pm 86.54$ & $104.91 \pm 27.45$ \\
DreamerV3-12M              & 10.495M           & $\mathbf{607.22 \pm 9.95}$ & $\mathbf{851.17 \pm 69.61}$ \\
VIBES (Ours)               & 3.257M            & $\underline{561.61 \pm 68.96}$ & $\underline{757.19 \pm 76.31}$ \\[1.5pt]
\midrule
No Adv. Obj.               & 1.386M            & $\underline{509.30 \pm 81.46}$ & $364.50 \pm 261.34$ \\[1.5pt]
\hline
\end{tabular}
}

\caption{Model capacity and performance on Realistic ManiSkill at 100K training steps. Values are mean $\pm$ 95\% confidence intervals across 6 seeds (3 for TIA and Denoised MDP). $\dagger$ denotes results taken from HRSSM \citep{hrssm}, converted from mean $\pm$ SD to 95\% confidence intervals. Bolded numbers indicate the highest mean return in each column.
Underlined numbers are not significantly different from the bolded entry (Welch's t-test, $p > 0.05$). TIA's underline on Lift Cube reflects high variance across only 3 seeds. No Adv. Obj. is our ablation without the adversarial term.}
\label{tab:realistic-maniskill}
\end{table}

 VIBES matches DreamerV3-12M on both tasks (no significant difference; Welch's t-test, $p = 0.25$ on Lift Cube and $p = 0.11$ on Turn Faucet) while using roughly $3\times$ fewer learned parameters (3.3M vs. 10.5M). It significantly outperforms all other baselines on both tasks, except TIA on Lift Cube, where TIA's 3-seed variance is too high to resolve a difference. Removing the adversarial objective (No Adv.\ Obj.) has no significant effect on Lift Cube ($p = 0.36$), consistent with our hypothesis in Section~\ref{sec:dcs} that the adversarial term matters least when the task-relevant state is fully observable: Realistic ManiSkill's static backgrounds leave the task state largely observable. On Turn Faucet, the no-adversary ablation has a significantly lower return ($p = 0.03$) and a much wider confidence interval, because two of its six seeds fail to learn the task. This suggests that even when it contributes little at convergence, the adversary supplies the encoder with useful gradient early in training, making learning more reliable.

\subsection{High-Dimensional Continuous Control from Vector States}
\label{sec:vector-states}

\begin{figure}[t]
  \centering
  \includegraphics[width=0.5\linewidth]{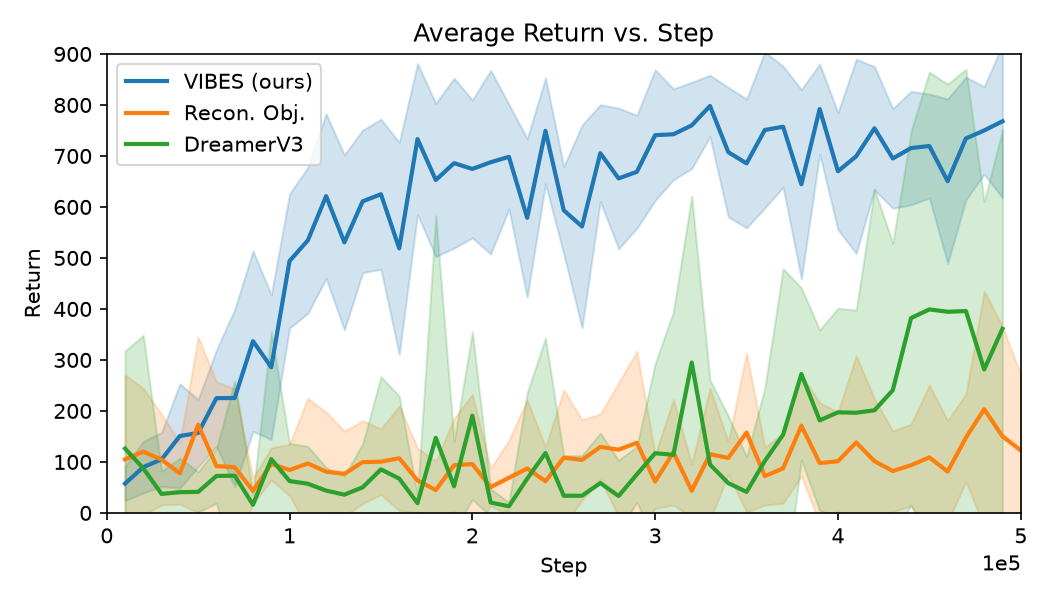}
  \caption{\textbf{100-link swimmer (vector states): learning curves.}
  Mean episode return over 4 runs versus environment steps, with 95\% confidence intervals shaded, for VIBES, Recon.\ Obj.\ (Section~\ref{sec:dcs}), and DreamerV3 on the 100-link swimmer task with 401-dimensional vector observations. VIBES learns faster and reaches higher returns than both reconstruction-based baselines.}
  \label{fig:swimmer-sidecap}
\end{figure}

To test generality beyond image-based tasks, we extend the \emph{Swimmer} environment from the DeepMind Control Suite \citep{dmc} to a 100-link planar snake robot with a 204-dimensional state and a 401-dimensional observation---a high-dimensional, vector-based control task. The agent must propel its nose to a fixed target, with a reward that decays smoothly with distance according to a Lorentzian function. Rather than controlling all 99 joints directly, we parameterize the action space with a spatiotemporal sinusoid that maps four parameters to every joint torque. Implementation details are in Appendix~\ref{app:swimmer}.

This setting tests whether VIBES scales to high-dimensional vector states, where reconstruction is difficult, and whether it works without the image-specific augmentations, such as random crops and color jittering, that prior approaches rely on.

We compare VIBES against DreamerV3 and against Recon.\ Obj., the reconstruction ablation from Section~\ref{sec:dcs}. Across 4 runs, VIBES learns faster and more stable policies and achieves higher returns than both baselines (Figure~\ref{fig:swimmer-sidecap}). Recon.\ Obj.\ struggles to make progress in the high-dimensional state space, and DreamerV3 improves only late in training, with high variance across runs. These results reinforce our claim that bisimulation-based objectives offer a principled and general approach to state abstraction, applicable to both visual and non-visual domains.

\subsection{Latent Space Analysis}
\label{sec:latent-analysis}

\begin{figure}[tb]
  \centering
  \begin{subfigure}{0.40\textwidth}
    \centering
    \includegraphics[width=\linewidth]{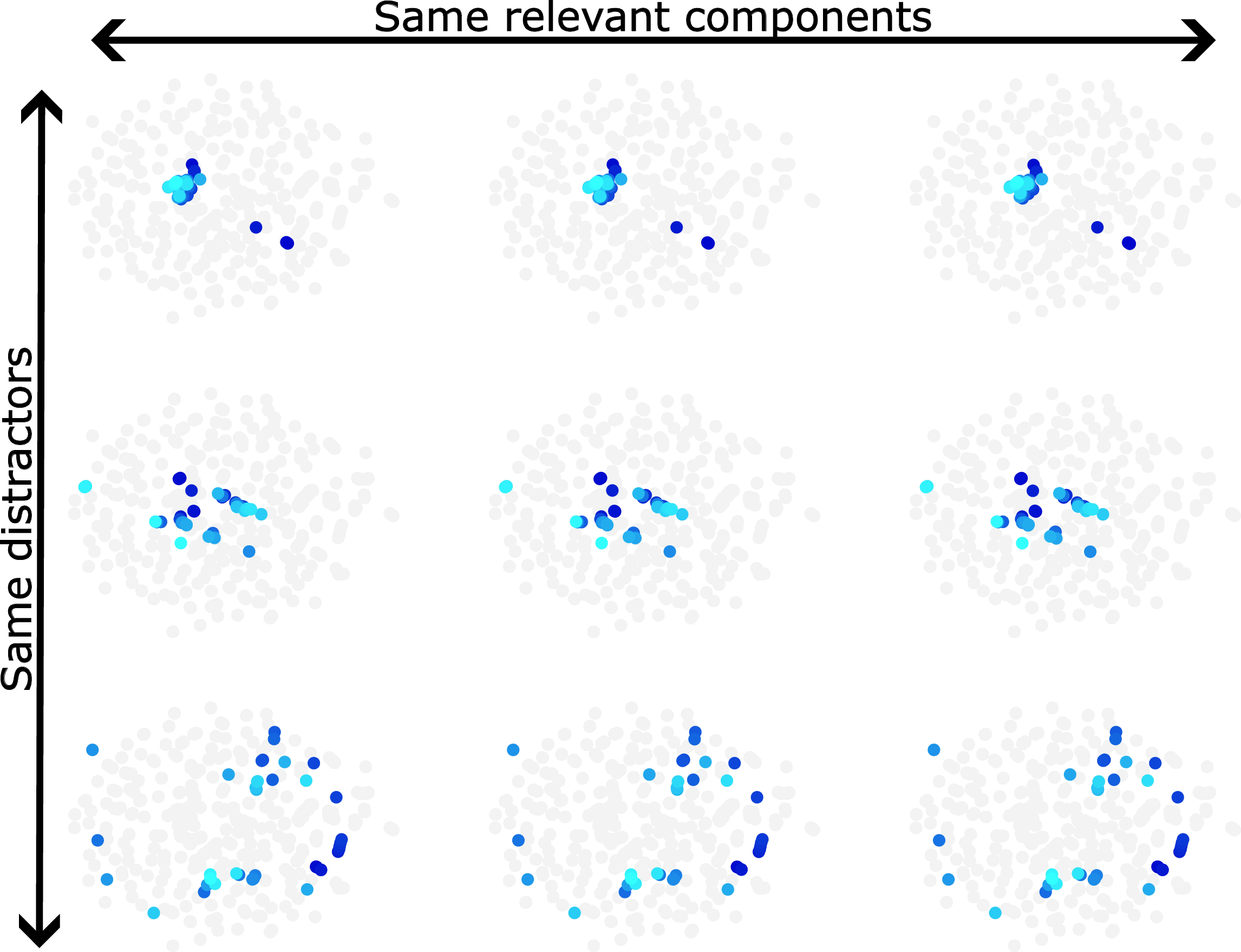}
    \caption{UMAP embeddings (3$\times$3).}
    \label{fig:umap-array}
  \end{subfigure}\hfill
  \begin{subfigure}{0.40\textwidth}
    \centering
    \includegraphics[width=\linewidth]{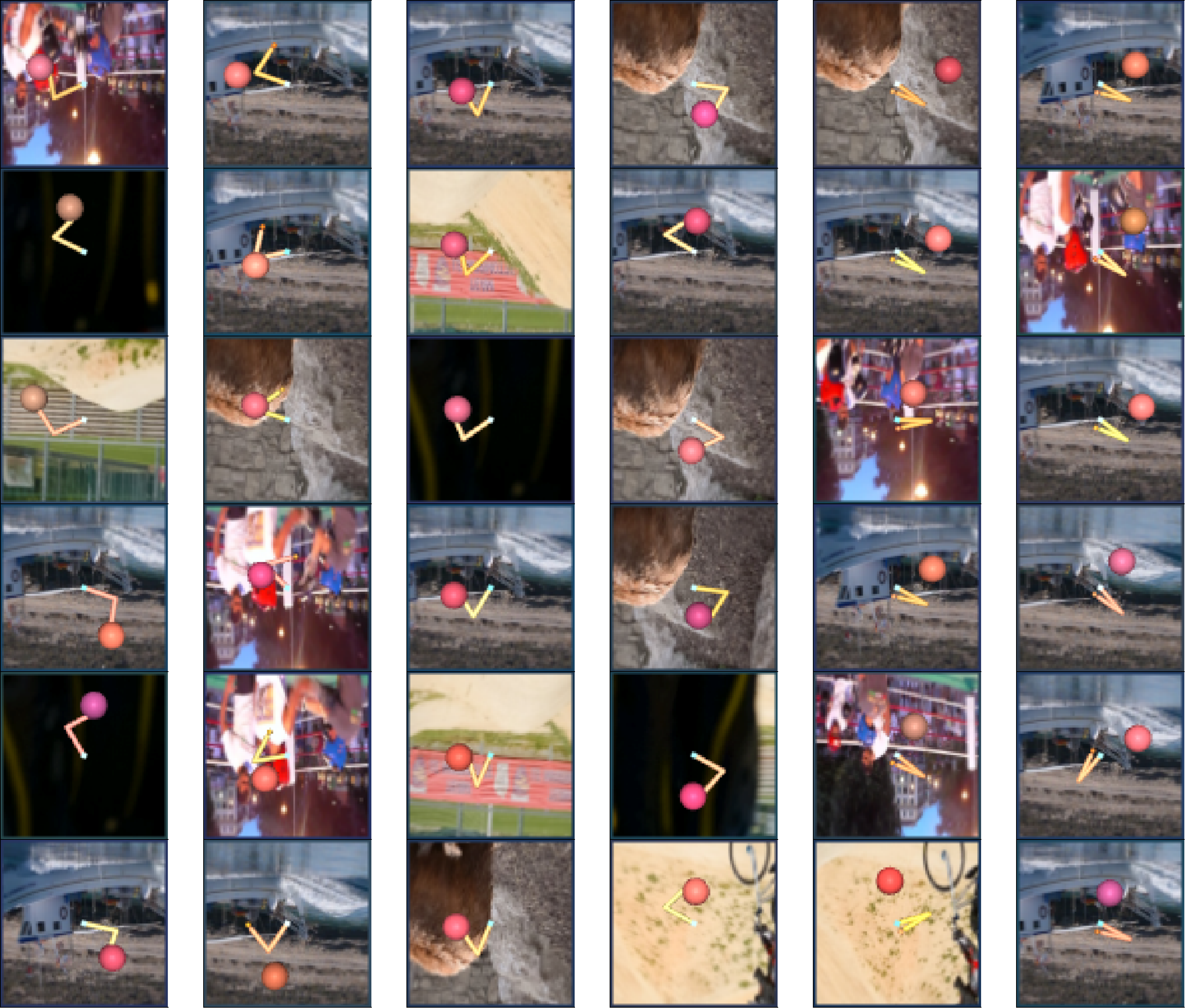}
    \caption{Nearest neighbors in latent space.}
    \label{fig:latent-nn}
  \end{subfigure}

  \captionsetup{width=.98\linewidth}
  \caption{\textbf{Latent-space analyses on \emph{Reacher-easy}.} (a) UMAP projections of 9 encoded trajectories in a $3 \times 3$ grid; rows share task states and columns share background distractors. Latent states change with the task state but not with the background. (b) Five nearest latent neighbors for six query observations (top of each column). Neighbors differ in background but match the query's arm pose relative to the goal, sometimes rotated about the origin, showing that the representation is task-relevant and invariant to joint rotations of the arm and goal.}
  \label{fig:latent-side-by-side}
\end{figure}

We generate 10 \emph{Reacher-easy} trajectories from distinct initial states and render each over 10 background videos, producing 100 unique trajectory-background pairs. Each trajectory is a sequence of 100 states (the agent's joint angles and velocities and the goal position). We encode the frames of every pair and map the resulting latent states to two dimensions with UMAP \citep{umap}. Figure~\ref{fig:umap-array} shows a $3 \times 3$ subset. Each panel displays the latent states from all 100 pairs in gray and highlights those of one pair in blue. Moving across a row varies the background while keeping the trajectory fixed; moving down a column varies the trajectory while keeping the background fixed. The blue points barely move within a row but shift across rows, indicating that the representation varies with task state and is largely invariant to background distractors.

The UMAP analysis does not reveal which aspects of the observation are preserved versus collapsed, so we also conduct a nearest-neighbor analysis. We encode the dataset collected during online training on \emph{Reacher-easy}, select 20 random latent states, and retrieve each one's five nearest neighbors from other trajectories under standardized L2 distance, which normalizes each dimension by its variance. Figure~\ref{fig:latent-nn} shows six representative queries.

The analysis supports two main findings. First, as expected, the encoder is largely invariant to distractors: states with different backgrounds and agent colors are grouped together in latent space. Second, the encoder appears to have learned a rotational symmetry of the Reacher problem: because the optimal action sequence is unchanged by jointly rotating the arm and goal location about the origin, the encoder clusters together configurations that are equivalent under such rotations. This suggests that the learned representation is even more compact than the true underlying vector state, which preserves absolute angles. These results indicate that the encoder captures task-relevant structure while discarding both distractors and redundant symmetries, consistent with the bisimulation perspective.

\begin{figure}[t]
  \centering
  \begin{subfigure}{0.48\textwidth}
    \centering
    \includegraphics[width=\linewidth]{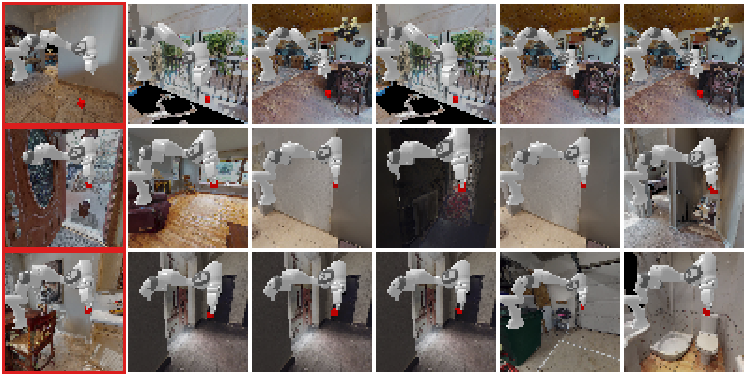}
    \caption{VIBES.}
    \label{fig:maniskill-nn-vibes}
  \end{subfigure}\hfill
  \begin{subfigure}{0.48\textwidth}
    \centering
    \includegraphics[width=\linewidth]{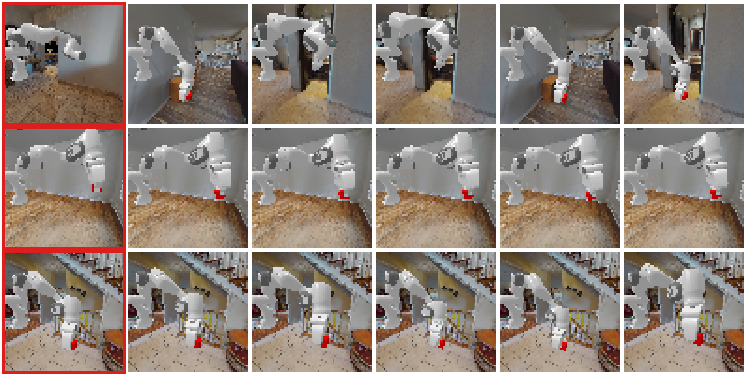}
    \caption{DreamerV3-12M.}
    \label{fig:maniskill-nn-dreamer}
  \end{subfigure}

  \captionsetup{width=.98\linewidth}
  \caption{\textbf{Nearest latent neighbors on Lift Cube (query outlined in red).} (a) VIBES: neighbors come from different scenes but share the query's gripper and cube configuration. (b) DreamerV3-12M: neighbors share the query's scene, even when gripper and cube configuration differ (top row).}
  \label{fig:maniskill-nn}
\end{figure}

We repeat the nearest-neighbor analysis on Realistic ManiSkill. For each of 6 seeds, we encode 50 episodes (5,000 frames) and retrieve each frame's five nearest neighbors under the same standardized L2 distance, excluding frames from the query's own episode so that adjacent, near-identical frames do not dominate the result. Figure~\ref{fig:maniskill-nn} shows the difference: DreamerV3-12M organizes its latent space by background, while VIBES organizes it by task state. To quantify this, we measure the proportion of neighbors that share the query's background scene. Across six seeds (mean $\pm$ 95\% confidence interval), this is $8.0 \pm 3.3\%$ for VIBES versus $40.9 \pm 7.1\%$ for DreamerV3-12M on Lift Cube, and $5.6 \pm 2.4\%$ versus $41.0 \pm 7.1\%$ on Turn Faucet. An encoder with no dependence on the background would score about $1/88 \approx 1.1\%$, since scenes are drawn from a pool of 88. This suggests that VIBES performs comparably to DreamerV3-12M with a third of the parameters (Table~\ref{tab:realistic-maniskill}) partly because it does not spend capacity on background distractors. Additional latent-space visualizations are provided in Appendix~\ref{app:latent-analysis}.

\section{Conclusions}
We introduced \textbf{VIBES} (Variational Inference for Bisimulation-based Encoded States), a model-learning objective that replaces pixel reconstruction with two control-aligned terms: a \emph{reward sufficiency} term and a \emph{dynamics sufficiency} term. We showed that the VIBES objective decomposes into the encoder's violation of reward and dynamics sufficiency, so that whenever exact sufficiency is attainable, its global optima induce bisimulation relations on histories . A fixed-capacity latent bottleneck then encourages the representation to collapse distractors. Empirically, VIBES serves as a drop-in replacement for Dreamer’s model-learning component and achieves state-of-the-art performance on the Distracting Control Suite without image-specific augmentations. On Realistic ManiSkill, VIBES extends this robustness to robotic manipulation with realistic background distractors, showing no significant difference from DreamerV3-12M (Welch's t-test) with a third of its learned parameters. We further demonstrated applicability to high-dimensional vector-state control, and latent-space analyses showed that the learned latents are invariant to distractors, capture symmetries of the dynamics, and, unlike the reconstruction-based DreamerV3-12M, are organized by task-relevant state rather than background scene. Overall, VIBES advances representation learning for MBRL by bridging the gap between the theoretically appealing framework of bisimulation relations and the empirically successful and robust framework of VI.

\section{Limitations and Future Work}
\begin{enumerate}[leftmargin=1.5em]
    \item \textbf{Approximation to bisimulation.} Theorem~\ref{thm:exact} requires exact sufficiency, which our fixed-variance encoder cannot attain (Proposition~\ref{prop:fixed-sigma}). Theorem~\ref{thm:approx} instead characterizes the objective through $\Delta(\phi)$, which cannot be estimated directly and which we do not translate into a bound on value loss (App.~\ref{app:fixed-sigma}).
 \item \textbf{Choice of abstraction.} Bisimulation is not the only task-aligned notion of state abstraction. Alternatives—e.g., value-equivalence, successor features, or predictive state representations—may be preferable for certain objectives (e.g., policy transfer or exploration). Extending VIBES to interpolate among these notions is an open direction.
    \item \textbf{Computation.} The adversary adds compute and memory overhead compared to pure reconstruction-free objectives.
    \item \textbf{Generalizability.} VIBES learns an abstraction relative to a single reward function, discarding information that doesn't influence the reward. This yields distractor robustness at small model scale, but limits transfer: information discarded for one task may be essential for another, so the representation must be relearned when the reward changes. Large-scale world models such as DreamerV4~\citep{dreamerv4} make the opposite trade-off, pretraining task-agnostic representations on unlabeled video at the cost of substantially larger models and datasets.
    \item \textbf{Scope of evaluation.} We evaluate image-based control with distractors, robotic manipulation, and vector-state control in simulation. Evaluation on physical robots is outside this scope.
    \item \textbf{Design choices in the bottleneck.} Our fixed-variance, unit-norm latent parameterization imposes a specific information bottleneck that may interact with task difficulty. Learning the bottleneck (e.g., via learnable variance or rate–distortion trade-offs) could improve adaptability across domains. The pressure to discard distractors comes from this bottleneck rather than the objective (Section~\ref{sec:model-objective}); our theory does not formalize this.
\end{enumerate}

\subsubsection*{Reproducibility Statement}
We have made efforts to ensure that our results are reproducible. Appendix~\ref{app:implementation} contains the full implementation details: the model architecture (\ref{app:model_arch}), the training algorithm (\ref{app:model_training}), all hyperparameters (\ref{app:hyperparams}), and the environment configurations for the Distracting Control Suite (\ref{app:dcs_config}), Realistic ManiSkill (\ref{app:maniskill-protocol}), and the 100-link swimmer (\ref{app:swimmer}). All reported results include the number of seeds and 95\% confidence intervals; Appendix~\ref{app:significance-tests} reports Welch's $t$-tests comparing VIBES with every method, and Appendix~\ref{app:dcs-learning-curves} shows learning curves for the Distracting Control Suite tasks.

\bibliography{iclr2026_conference,ben_edits_refs}
\bibliographystyle{iclr2027_conference}

\appendix
\raggedbottom

\section{Derivation of the Model Objective}
\label{app:model_obj_derivation}

\paragraph{Notation} Throughout the derivation, we use \(\vec{o}\), \(\vec{a}\), \(\vec{r}\), and \(\vec{\tilde{s}}\) to denote temporal sequences of observations, actions, rewards, and state representations, respectively. We use \(p(\cdot)\) to denote \textit{true} probability distributions (\textit{i.e., those specified by the environment}) and \(\hat{p}(\cdot)\) to denote approximations to said true distributions (\textit{i.e.}, those specified by our model).  We use \(\tilde{s}_t \sim e_{\phi}(\cdot|o_{\leq t},a_{<t})\) to denote the distribution over the state representation at time \(t\), which depends on the observations before and including time \(t\), and the actions up to time \(t\), and is parameterized by \(\phi\).  Note that we do not place a circumflex (\(\quad \hat{} \quad \)) over \(e_{\phi}\), because this is our \textit{chosen} state representation and does not represent an approximation to some underlying true distribution.

As in PlaNet \citep{planet} and Dreamer \citep{dreamer}, we form a variational lower bound by marginalizing over the latent sequence and applying Jensen's inequality to the logarithm. This gives a lower bound on the expected log ratio of the approximate and true distributions over our observations and rewards given actions,

\begin{equation}
    \label{eq:lb}
    \mathbb{E}_{\vec{o},\vec{r},\vec{a} \sim D} \left[ \log \frac{\hat{p}(\vec{o},\vec{r}|\vec{a})}{p(\vec{o},\vec{r}|\vec{a})} \right] \geq  \mathbb{E}_{\vec{o},\vec{r},\vec{a} \sim D} \mathbb{E}_{\vec{\tilde{s}} \sim e_{\phi}} \left[ \log \frac{\hat{p}(\vec{o},\vec{r},\vec{\tilde{s}}|\vec{a})}{e_{\phi}(\vec{\tilde{s}}|\vec{o},\vec{a})p(\vec{o},\vec{r}|\vec{a})}\right].
\end{equation}

From here, we diverge from the Dreamer derivation, factoring both the numerator and denominator differently.

\paragraph{Numerator} In the Dreamer derivation, \(\hat{p}(\vec{o},\vec{r},\vec{\tilde{s}}|\vec{a})\) factors into an \textit{observation decoder} \(\hat{p}_{\theta}(o_{t}|\tilde{s}_{t})\), a \textit{latent dynamics model} \(\hat{p}_{\theta}(\tilde{s}_{t}|\tilde{s}_{t-1},a_{t-1})\), and a \textit{reward decoder} \(\hat{p}_{\theta}(r_t|\tilde{s}_t,a_t)\).  We instead factor the numerator according to

\begin{equation}
    \label{eq:num}
    \hat{p}(\vec{o},\vec{r},\vec{\tilde{s}}|\vec{a}) = \prod_{t=1}^{T} \textcolor{red}{p(o_{t}|\tilde{s}_{ \leq t}, o_{< t}, a_{< t})} \hat{p}_{\theta}(\tilde{s}_{t}|\tilde{s}_{t-1},a_{t}) \hat{p}_{\theta}(r_t|\tilde{s}_t,a_t).
\end{equation}

Note that the term in \textcolor{red}{red} has two key differences from the corresponding term in Dreamer.  First, it conditions on \(o_{< t}\), thereby avoiding the assumption that the system is Markovian in \(s\).  This allows us to avoid learning a state representation that is incentivized to encode all (potentially irrelevant) information from past observations that is predictive for the current observation (see Figure~\ref{fig:pgm-side-by-side} right for a visual representation of these assumptions).  Second, we use the \textit{true} (but unknown) distribution over \(p(o_{t}|\tilde{s}_{\le t}, o_{< t}, a_{< t})\) rather than a model-approximated distribution, allowing us to avoid having to learn an observation decoder and the associated challenges of representing high-dimensional probability distributions.  As we shall see, the fact that this distribution is unknown will not impact our ability to optimize the resulting model objective because this term will cancel.

\paragraph{Denominator} In the Dreamer derivation, \(p(\vec{o},\vec{r}|\vec{a})\) is treated as a constant and eliminated.  We instead chose to group it together with the encoder distribution term and re-factor it.  

We recognize that the denominator \(e_{\phi}(\vec{\tilde{s}}|\vec{o},\vec{a})\, p(\vec{o},\vec{r}|\vec{a})\) is the \textit{true} joint distribution over \(\vec{o}\), \(\vec{r}\), \(\vec{\tilde{s}}\) given \(\vec{a}\). This is because \(e_{\phi}\) describes our chosen state representation, not an approximation to some underlying distribution over latent states, so \(p(\vec{\tilde{s}}|\vec{o},\vec{a}) = e_{\phi}(\vec{\tilde{s}}|\vec{o},\vec{a})\). Since the encoder never observes rewards, conditioning on \(\vec{r}\) does not change this. Therefore

\begin{equation}
\label{eq:joint-true}
p(\vec{o},\vec{r},\vec{\tilde{s}}|\vec{a}) = p(\vec{\tilde{s}}|\vec{o},\vec{r},\vec{a})\, p(\vec{o},\vec{r}|\vec{a}) = e_{\phi}(\vec{\tilde{s}}|\vec{o},\vec{a})\, p(\vec{o},\vec{r}|\vec{a}).
\end{equation}

We factor this joint in three steps. First, group variables by timestep:

\begin{equation}
\label{eq:group-by-timestep}
p(\vec{o},\vec{r},\vec{\tilde{s}}|\vec{a}) = \prod_{t=1}^{T} p(o_t, r_t, \tilde{s}_t \mid o_{<t}, r_{<t}, \tilde{s}_{<t}, a_{\leq t}).
\end{equation}

Second, apply the chain rule within each timestep:

\begin{equation}
\label{eq:chain-rule}
= \prod_{t=1}^{T} p(\tilde{s}_t \mid \tilde{s}_{<t}, o_{<t}, r_{<t}, a_{<t})\, p(o_t \mid \tilde{s}_{\leq t}, o_{<t}, r_{<t}, a_{<t})\, p(r_t \mid \tilde{s}_{\leq t}, o_{\leq t}, r_{<t}, a_{\leq t}).
\end{equation}
Third, we drop redundant conditioning. Observations and state representations are conditionally independent of past rewards given past observations and actions. Because $\tilde{s}_{<t}$ is computed from $o_{<t}$ and $a_{<t}$ (plus independent encoder noise), $r_t$ is conditionally independent of $\tilde{s}_{\leq t}$ given $o_{\leq t}$ and $a_{\leq t}$. This gives us

\begin{equation}
    \label{eq:denom}
    p(\vec{o},\vec{r},\vec{\tilde{s}}|\vec{a}) =   \prod_{t=1}^{T} p(o_{t}|\tilde{s}_{ \leq t}, o_{< t}, a_{< t}) p(\tilde{s}_{t}| \tilde{s}_{<t}, o_{< t},a_{< t}) p(r_t|o_{\leq t}, a_{\leq t}).
\end{equation}

Substituting \eqref{eq:num} and \eqref{eq:denom} back into \eqref{eq:lb} gives us

\begin{align}
     \mathbb{E} \left[ \log \frac{\hat{p}(\vec{o},\vec{r}|\vec{a})}{p(\vec{o},\vec{r}|\vec{a})} \right] & \geq \mathbb{E} \left[ \log \frac{ \prod_{t=1}^{T} \cancel{p(o_{t}|\tilde{s}_{ \leq t}, o_{< t}, a_{\leq t})} \hat{p}_{\theta}(\tilde{s}_{t}|\tilde{s}_{t-1},a_{t-1}) \hat{p}_{\theta}(r_t|\tilde{s}_t,a_t)}{  \prod_{t=1}^{T} \cancel{p(o_{t}|\tilde{s}_{ \leq t}, o_{< t}, a_{< t})} p(\tilde{s}_{t}| \tilde{s}_{< t}, o_{< t},a_{< t}) p(r_t|o_{\leq t}, a_{\leq t})}
     \right] \\
     & = \mathbb{E} \left[ \sum \log \frac{\hat{p}_{\theta}(\tilde{s}_{t}|\tilde{s}_{t-1},a_{t-1})}{p(\tilde{s}_{t}| \tilde{s}_{< t}, o_{< t},a_{< t})}  +  \log \hat{p}_{\theta}(r_t|\tilde{s}_t,a_t) \right] + \textcolor{gray}{\mathcal{H}(p(\vec{r}|\vec{o},\vec{a}))}. \label{eq:model_obj_before_adversary_net}
\end{align}

Because \(\textcolor{gray}{\mathcal{H}(p(\vec{r}|\vec{o},\vec{a}))}\) does not depend on our model parameters, it can be treated as a constant and ignored. \(p(\tilde{s}_{t}| \tilde{s}_{< t}, o_{< t},a_{< t})\) is unknown; however, if the class \(\Psi\) of adversaries contains it (an assumption we maintain throughout Appendix~\ref{app:proofs}), then by Gibbs' inequality it is given by

\begin{equation}
\label{eq:psi_objective}
p(\tilde{s}_{t}| \tilde{s}_{< t}, o_{< t},a_{< t}) = \hat{p}_{\psi*}(\tilde{s}_{t}| \tilde{s}_{< t}, o_{< t},a_{< t}) = \underset{\psi \in \Psi}{\mathrm{argmax}}
\mathbb{E} \left[ \log \hat{p}_{\psi}(\tilde{s}_{t}|\tilde{s}_{< t}, o_{< t},a_{< t} ) \right].
\end{equation}

Substituting this term into \eqref{eq:model_obj_before_adversary_net} yields our final model objective,

\begin{equation}
    \label{eq:final-objective}
    \mathcal{L}(\theta,\phi) = \mathbb{E} \left[ \sum \log \frac{\hat{p}_{\theta}(\tilde{s}_{t}|\tilde{s}_{t-1},a_{t-1})}{\hat{p}_{\psi*}(\tilde{s}_{t}| \tilde{s}_{< t}, o_{< t},a_{< t})}  +  \log \hat{p}_{\theta}(r_t|\tilde{s}_t,a_t) \right].
\end{equation}

We refer to $\hat{p}_{\psi*}$ as the \textit{adversary network} because optimizing it \textit{reduces} $\mathcal{L}$, placing $\psi$ in an adversarial game with $\phi$ and $\theta$. Similarly, we refer to 
\(
\mathbb{E}\left[\sum_t \log \tfrac{1}{\hat{p}_{\psi*}(\tilde{s}_{t} \mid \tilde{s}_{< t}, o_{< t}, a_{< t})}\right]
\)
as the \textit{adversarial objective component}. In practice, to ensure that the adversary network never underperforms the dynamics model in predicting $\tilde{s}_t$, we clip each per-dimension term
\(
\mathbb{E}\!\left[ \log \tfrac{\hat{p}_{\theta}(\tilde{s}^{(i)}_{t} \mid \tilde{s}_{t-1}, a_{t-1})}{\hat{p}_{\psi*}(\tilde{s}^{(i)}_{t} \mid \tilde{s}_{< t}, o_{< t}, a_{< t})} \right]
\)
to be non-positive. This clipping is a stabilization heuristic for when the adversary is slow to converge: it departs from \eqref{eq:final-objective} whenever it is active, and our theoretical results (Theorems~\ref{thm:exact}--\ref{thm:approx} and Proposition~\ref{prop:envelope}) are stated for the unclipped objective.

\section{Proofs}
\label{app:proofs}
\subsection{\texorpdfstring{Proof of Lemma \ref{lem:sufficiency}}{Proof of Lemma 3.1}}
\label{app:lemma_proof}

Throughout, a history is $H_t = (o_{\le t}, a_{<t})$. After taking action $a$ in history $H$, the next observation is drawn as $o' \sim P(\cdot\mid H,a)$, giving the next history $H' = (H,a,o')$, and the next latent is drawn as $\tilde{s}' \sim e_\phi(\cdot\mid H')$ with fresh encoder noise. Since the environment does not observe encoder noise, the next latent is independent of the current latent given $(H,a)$:
\begin{equation}
P(\tilde{s}'\mid H,\tilde{s}=s,a) = \int e_\phi(\cdot\mid H')\,P(dH'\mid H,a) =: Q_\phi(\cdot\mid H,a) \quad\text{for every } s,
\label{eq:next-latent-mixture}
\end{equation}
and similarly $P(r\mid H,\tilde{s}=s,a) = R(\cdot\mid H,a)$, the environment's reward distribution. (For the augmented latent state of App.~\ref{app:augmented}, the next latent does depend on the current one, which is why Lemma~\ref{lem:sufficiency} conditions on $\tilde{s}=s$.) Each equivalence class $\tilde{H}$ of $\sim_{e_\phi}$ corresponds to a single encoder distribution.

\paragraph{Identifiability.} We call the encoder family \emph{identifiable} if, for any two distributions $\nu,\nu'$ over histories,
\[
\int e_\phi(\cdot\mid H)\,\nu(dH) = \int e_\phi(\cdot\mid H)\,\nu'(dH)
\quad\Longrightarrow\quad
\nu(\tilde{H}) = \nu'(\tilde{H})\quad \forall \tilde{H}\in\tilde{\mathcal{H}},
\]
that is, the distribution of a latent determines the distribution over the equivalence classes it was drawn from. This holds trivially for deterministic encoders, since each latent identifies its class. It also holds for Gaussian encoders with fixed covariance, $e_\phi(\cdot\mid H) = \mathcal{N}(\mu_\phi(H),\sigma^2 I)$: classes correspond to values of $\mu_\phi$, the mixture is the distribution of $\mu_\phi(H)$ convolved with $\mathcal{N}(0,\sigma^2 I)$, and convolution with a Gaussian is injective on probability measures because the Gaussian characteristic function has no zeros.

\paragraph{Proof of Lemma~\ref{lem:sufficiency}.}
Let $H^i \sim_{e_\phi} H^j$, so $e_\phi(\cdot\mid H^i) = e_\phi(\cdot\mid H^j)$, and in particular the two distributions have the same support. Fix any $s$ in this support and any action $a$.

\emph{Rewards.} By \eqref{eq:rew-suff}, $R(\cdot\mid H^i,a) = \rho(\cdot\mid s,a) = R(\cdot\mid H^j,a)$. A deterministic reward function $R(H,a)$ is the special case $R(\cdot\mid H,a) = \delta(\cdot - R(H,a))$, so this generalizes the reward condition of \eqref{eq:bisim-def} to stochastic rewards.

\emph{Transitions.} By \eqref{eq:dyn-suff} and \eqref{eq:next-latent-mixture}, $Q_\phi(\cdot\mid H^i,a) = \kappa(\cdot\mid s,a) = Q_\phi(\cdot\mid H^j,a)$. Thus the latent mixtures generated by $P(\cdot\mid H^i,a)$ and $P(\cdot\mid H^j,a)$ coincide, and by identifiability these two distributions assign equal probability to every equivalence class $\tilde{H}'$, i.e., $P_{\tilde{H}'\mid H,a}(\tilde{H}'\mid H^i,a) = P_{\tilde{H}'\mid H,a}(\tilde{H}'\mid H^j,a)$.

Both conditions of \eqref{eq:bisim-def} hold, so $\sim_{e_\phi}$ is a bisimulation relation. \qed

\subsection{\texorpdfstring{Proofs of Theorems \ref{thm:exact} and \ref{thm:approx}}{Proofs of Theorems 3.2 and 3.3}}
\label{app:theorem_proof}

\paragraph{Proof overview.} After adding and subtracting $\mathbb{E}[\log p(r_t\mid o_{\le t},a_{\le t})]$, each term of $\mathcal{L}$ is the negative expected log ratio between a history-conditioned true conditional and a latent-conditioned model. The chain rule for mutual information (Lemma~\ref{lem:decomp}) splits each into a conditional mutual information term (what the history adds beyond the latent) and a KL term (model misfit), giving Theorem~\ref{thm:approx}. Under joint realizability, both vanish at every global optimum, and Lemma~\ref{lem:sufficiency} yields a bisimulation (Theorem~\ref{thm:exact}).

\begin{lemma}[Chain-rule decomposition]
\label{lem:decomp}
Let $(Y,P,Z)$ be random variables with $Y \perp\!\!\!\perp Z \mid P$, and let $q(y\mid z)$ be a conditional density. Then
\[
\mathbb{E}\big[\log p(Y\mid P) - \log q(Y\mid Z)\big] = I(Y;P\mid Z) + \mathbb{E}_Z\big[D_{\mathrm{KL}}\big(p(Y\mid Z)\,\|\,q(Y\mid Z)\big)\big],
\]
where both sides may be $+\infty$.
\end{lemma}
\textit{Proof.} Conditional independence gives $p(Y\mid P) = p(Y\mid P,Z)$. Writing $\log p(Y\mid P,Z) - \log q(Y\mid Z) = [\log p(Y\mid P,Z) - \log p(Y\mid Z)] + [\log p(Y\mid Z) - \log q(Y\mid Z)]$ and taking expectations, the first bracket gives $I(Y;P\mid Z)$ and the second gives the expected KL divergence. \qed

\paragraph{Proof of Theorem~\ref{thm:approx}.}
With the adversary equal to the true conditional, \eqref{eq:model_loss} reads
\[
\mathcal{L}(\theta,\phi) = \sum_t \mathbb{E}\Big[\log\hat{p}_\theta(\tilde{s}_t\mid Z^s_t) - \log p(\tilde{s}_t\mid P^s_t) + \log\hat{p}_\theta(r_t\mid Z^r_t)\Big],
\]
where $P^s_t = (\tilde{s}_{<t},o_{<t},a_{<t})$, $Z^s_t = (\tilde{s}_{t-1},a_{t-1})$, $P^r_t = (o_{\le t},a_{\le t})$, and $Z^r_t = (\tilde{s}_t,a_t)$. Adding and subtracting $C = \sum_t\mathbb{E}[\log p(r_t\mid P^r_t)]$, which is determined by the environment and the data distribution and so does not depend on $(\theta,\phi)$, gives
\[
\mathcal{L}(\theta,\phi) = C - \sum_t\mathbb{E}\big[\log p(\tilde{s}_t\mid P^s_t) - \log\hat{p}_\theta(\tilde{s}_t\mid Z^s_t)\big] - \sum_t\mathbb{E}\big[\log p(r_t\mid P^r_t) - \log\hat{p}_\theta(r_t\mid Z^r_t)\big].
\]
We apply Lemma~\ref{lem:decomp} to each sum. For the dynamics term, $Z^s_t$ is a function of $P^s_t$, so $\tilde{s}_t \perp\!\!\!\perp Z^s_t \mid P^s_t$ holds trivially. For the reward term, $r_t \perp\!\!\!\perp \tilde{s}_t \mid P^r_t$ because $\tilde{s}_t$ depends on the history only through $(o_{\le t},a_{<t})$ and encoder noise that the environment does not observe; this holds even when $a_t$ is chosen by a policy acting on $\tilde{s}_t$, since $a_t$ is included in $P^r_t$. The lemma yields $\mathcal{L} = C - \Delta(\phi) - M(\theta,\phi)$ with
\[
M(\theta,\phi) = \sum_t\mathbb{E}\Big[D_{\mathrm{KL}}\big(p(\tilde{s}_t\mid Z^s_t)\,\|\,\hat{p}_\theta(\tilde{s}_t\mid Z^s_t)\big) + D_{\mathrm{KL}}\big(p(r_t\mid Z^r_t)\,\|\,\hat{p}_\theta(r_t\mid Z^r_t)\big)\Big] \;\ge\; 0.
\]
Hence $\mathcal{L}\le C - \Delta(\phi)$, with equality iff $M=0$, i.e., iff $\hat{p}_\theta$ equals the best latent-conditioned predictors almost surely. If the model class contains these predictors for every $\phi$, then $\max_\theta\mathcal{L}(\theta,\phi) = C-\Delta(\phi)$. \qed

\paragraph{Proof of Theorem~\ref{thm:exact}.}
By Theorem~\ref{thm:approx}, $\mathcal{L}(\theta,\phi)\le C$ for all $(\theta,\phi)$. Under \eqref{eq:realizable}, each expected log ratio in the proof above vanishes at $(\theta^\circ,\phi^\circ)$, so $\mathcal{L}(\theta^\circ,\phi^\circ) = C$ and the maximum of $\mathcal{L}$ equals $C$. Any global maximizer $(\theta^*,\phi^*)$ therefore has
\[
\sum_t\mathbb{E}\Big[D_{\mathrm{KL}}\big(p(\tilde{s}_t\mid P^s_t)\,\|\,\hat{p}_{\theta^*}(\tilde{s}_t\mid Z^s_t)\big) + D_{\mathrm{KL}}\big(p(r_t\mid P^r_t)\,\|\,\hat{p}_{\theta^*}(r_t\mid Z^r_t)\big)\Big] = 0,
\]
so both KL divergences vanish almost surely, which is \eqref{eq:realizable} under $e_{\phi^*}$.

To connect these equalities to Lemma~\ref{lem:sufficiency}, note that earlier latents carry no information about the next latent beyond the history, the current latent, and the action, so $p(\tilde{s}_t\mid\tilde{s}_{<t},o_{<t},a_{<t}) = P(\tilde{s}_t\mid H_{t-1},\tilde{s}_{t-1},a_{t-1})$, and similarly $p(r_t\mid o_{\le t},a_{\le t}) = P(r_t\mid H_t,\tilde{s}_t,a_t)$. Equation~\ref{eq:realizable} therefore states that $P(\tilde{s}'\mid H,\tilde{s}=s,a) = \hat{p}_{\theta^*}(\tilde{s}'\mid s,a)$ and $P(r\mid H,\tilde{s}=s,a) = \hat{p}_{\theta^*}(r\mid s,a)$ almost surely over histories, actions, and latents. Under the coverage assumption, these hold for every history, action, and latent in the encoder's support, which are equations~\ref{eq:dyn-suff}--\ref{eq:rew-suff} with $\kappa=\rho=\hat{p}_{\theta^*}$. Lemma~\ref{lem:sufficiency} completes the proof. \qed

\subsection{\texorpdfstring{Gradients can be computed with the adversary held fixed}{Gradients can be computed with the adversary held fixed}}
\label{app:zero_derivative}
\label{app:envelope}

We justify treating $\psi^*$ as fixed when computing $\nabla_\theta\mathcal{L}$ and $\nabla_\phi\mathcal{L}$ (Section~\ref{sec:learning_bisim_relations}). Let $\omega = (\theta,\phi)$ collect the model and encoder parameters. Encoder latents are reparameterized as $\tilde{s}_t = \mu_\phi(o_{\le t},a_{<t}) + \sigma_e\epsilon_t$ with $\epsilon_t\sim\mathcal{N}(0,I)$, and expectations over $(\vec{o},\vec{r},\vec{a})$ are taken under the replay distribution $\mathcal{D}$, which does not depend on $\omega$ during a model update. Define the adversary's objective, with $\psi$ as a free argument,
\begin{equation}
F(\omega,\psi) := \mathbb{E}_{\mathcal{D},\vec{\epsilon}}\Big[\sum_t\log\hat{p}_\psi(\tilde{s}_t\mid\tilde{s}_{<t},o_{<t},a_{<t})\Big],
\end{equation}
its optimal value $V(\omega) := \max_{\psi\in\Psi}F(\omega,\psi)$, attained at $\psi^*(\omega)$ (\eqref{eq:psi_objective}), and the terms of the objective that do not involve the adversary, namely the dynamics and reward log-likelihoods,
\begin{equation}
A(\omega) := \mathbb{E}_{\mathcal{D},\vec{\epsilon}}\Big[\sum_t\log\hat{p}_\theta(\tilde{s}_t\mid\tilde{s}_{t-1},a_{t-1}) + \log\hat{p}_\theta(r_t\mid\tilde{s}_t,a_t)\Big].
\end{equation}
The model objective (\eqref{eq:final-objective}) is then $\mathcal{L}(\omega) = A(\omega) - V(\omega)$. Algorithm~\ref{alg:model_training} instead differentiates the surrogate $\tilde{\mathcal{L}}(\omega,\psi) := A(\omega) - F(\omega,\psi)$ with $\psi$ held fixed. In the Markov formulation of Section~\ref{sec:theory}, $F$ depends on $\omega$ only through $\phi$; we state the result for $\omega$ jointly because, with the augmented latent state of Appendix~\ref{app:augmented}, $F$ may also depend on $\theta$.

\begin{proposition}[Envelope property]
\label{prop:envelope}
Fix $\omega_0$ and let $\psi_0\in\arg\max_{\psi\in\Psi}F(\omega_0,\psi)$. If $V$ and $F(\cdot,\psi_0)$ are differentiable at $\omega_0$, then
\[
\nabla_\omega V(\omega_0) = \nabla_\omega F(\omega_0,\psi_0), \qquad\text{and consequently}\qquad \nabla_\omega\mathcal{L}(\omega_0) = \nabla_\omega\tilde{\mathcal{L}}(\omega_0,\psi_0).
\]
\end{proposition}
\textit{Proof.} Let $g(\omega) := V(\omega) - F(\omega,\psi_0)$. Since $V(\omega)\ge F(\omega,\psi)$ for every $\psi$, we have $g(\omega)\ge 0$ for all $\omega$, and $g(\omega_0)=0$ because $\psi_0$ attains the maximum at $\omega_0$. Hence $\omega_0$ is a global minimizer of $g$, and since $g$ is differentiable at $\omega_0$, $\nabla g(\omega_0)=0$, which is the first claim. The second follows because $\mathcal{L}(\omega)-\tilde{\mathcal{L}}(\omega,\psi_0) = -g(\omega)$. This argument follows \citet[Theorem~1]{milgrom2002envelope}. \qed

\paragraph{Which derivative is taken.} $\nabla_\omega F(\omega_0,\psi_0)$ is the full derivative of $F$ in its first argument with the adversary's parameters fixed at $\psi_0$. With $\psi_0$ fixed, $\phi$ enters $F$ only through the reparameterized latents, both as the adversary's prediction targets $\tilde{s}_t$ and as its inputs $\tilde{s}_{<t}$:
\[
\nabla_\phi F(\omega_0,\psi_0) = \mathbb{E}\Big[\sum_t\sum_{k\le t}\Big(\frac{\partial\tilde{s}_k}{\partial\phi}\Big)^{\!\top}\nabla_{\tilde{s}_k}\log\hat{p}_{\psi_0}(\tilde{s}_t\mid\tilde{s}_{<t},o_{<t},a_{<t})\Big].
\]
This quantity is generally nonzero. Proposition~\ref{prop:envelope} states only that the dependence of $\psi^*$ on $\omega$ contributes nothing to the gradient, so no derivative of $\psi^*(\omega)$ needs to be computed. In our implementation, the adversary's parameters are frozen during the model update while gradients propagate through its inputs and targets.

\paragraph{When $V$ is differentiable.} Sufficient conditions are given by Danskin's theorem~\citep{danskin1966theory}: $\Psi$ compact, $F$ and $\nabla_\omega F$ continuous in $(\omega,\psi)$, and $\nabla_\omega F(\omega,\psi)$ equal at every maximizer $\psi\in\arg\max F(\omega,\cdot)$. The last condition does not require a unique maximizing \emph{parameter}, which neural networks rarely have; it suffices that all maximizers induce the same conditional density $\hat{p}_\psi$, since $\nabla_\omega F$ depends on $\psi$ only through that density. This holds, for example, when $\Psi$ contains the true conditional $p(\tilde{s}_t\mid\tilde{s}_{<t},o_{<t},a_{<t})$, because the log score is strictly proper. The proposition itself does not require $\Psi$ to contain the true conditional; under that additional assumption, it recovers the property used by \citet[Proposition~2]{avb}.

\paragraph{Approximate adversaries.} In practice, $\psi$ only approximates $\psi^*$. If $\nabla_\omega F(\omega,\cdot)$ is $\beta$-Lipschitz in $\psi$, then $\|\nabla_\omega\tilde{\mathcal{L}}(\omega,\psi) - \nabla_\omega\mathcal{L}(\omega)\| \le \beta\,\|\psi-\psi^*(\omega)\|$, so the gradient error shrinks as the adversary tracks its optimum. This motivates the higher adversary learning rate in Algorithm~\ref{alg:model_training}. Appendix~\ref{app:adversary-gap} shows empirically that the adversary retains a predictive advantage over the dynamics model throughout training.

\paragraph{Clipping.} Proposition~\ref{prop:envelope} applies to the unclipped objective. The per-dimension clipping described in Appendix~\ref{app:model_obj_derivation} is a stabilization heuristic: when a clipped term is active, the adversary's first-order condition no longer cancels the corresponding contribution, because $\psi$ is shared across latent dimensions.

\subsection{Exact sufficiency with fixed-variance encoders}
\label{app:fixed-sigma}
This section explains why Theorem~\ref{thm:exact} does not apply to our implemented encoder, and why Theorem~\ref{thm:approx} does. In brief:
\begin{itemize}
    \item \textbf{Exact sufficiency is unattainable.} With a fixed-variance Gaussian encoder, exact reward sufficiency holds only if the reward does not depend on the history at all (Proposition~\ref{prop:fixed-sigma}). The joint realizability assumption of Theorem~\ref{thm:exact} therefore fails in any nontrivial task.
    \item \textbf{The obstruction is small.} It arises only from latent values that are possible under two histories with different rewards. When their encoder means are far apart relative to $\sigma_e = 0.01$, such values are extremely unlikely, so the encoder can come close to sufficiency.
    \item \textbf{Theorem~\ref{thm:approx} is the relevant guarantee.} Maximizing $\mathcal{L}$ minimizes the sufficiency violation $\Delta(\phi)$ within the encoder class. Proposition~\ref{prop:fixed-sigma} shows only that this minimum is strictly positive, not that it is large.
\end{itemize}

\begin{proposition}
\label{prop:fixed-sigma}
Let $e_\phi(\cdot\mid H) = \mathcal{N}(\mu_\phi(H),\sigma^2 I)$ with $\sigma>0$, or more generally let every encoder distribution have full support on the latent space. If reward sufficiency \eqref{eq:rew-suff} holds, then $R(\cdot\mid H,a) = R(\cdot\mid H',a)$ for all histories $H,H'$ and actions $a$.
\end{proposition}
\textit{Proof.} Fix any latent $s$. Because every encoder distribution has full support, $s$ lies in the support of both $e_\phi(\cdot\mid H)$ and $e_\phi(\cdot\mid H')$, so \eqref{eq:rew-suff} gives $R(\cdot\mid H,a) = \rho(\cdot\mid s,a) = R(\cdot\mid H',a)$. \qed

\paragraph{Intuition.} Gaussian noise has full support, so every latent value is possible, though perhaps extremely unlikely, under every history. Exact reward sufficiency requires the reward distribution given a latent value to be the same whichever history produced it. Applied to a latent value shared by two histories, this forces their reward distributions to coincide, and every pair of histories shares such a value. The same argument, applied to \eqref{eq:dyn-suff} and \eqref{eq:next-latent-mixture}, shows that exact dynamics sufficiency requires the distribution of the next latent not to depend on the history beyond the previous action. Together, these would leave the latent with no useful information.

\paragraph{Why this does not undermine the method.} Theorem~\ref{thm:exact} therefore describes an idealized limit, while Theorem~\ref{thm:approx} governs the implemented method: the encoder is driven to minimize $\Delta(\phi)$, whose minimum over the fixed-variance class is strictly positive but need not be large. This is consistent with the interpretation of the encoder as a finite-capacity channel (Section~\ref{sec:model-objective}): the noise that bounds the latent's information content, and so pressures the encoder to discard distractors, is the same noise that prevents any latent from being exactly sufficient.

\paragraph{Limits of this analysis.} $\Delta(\phi)$ cannot be estimated directly, since doing so requires the unknown reward entropy and an exact adversary. Translating it into a bound on value loss is future work: approximate-information-state results~\citep{subramanian2022approximate} suggest a route, but they bound worst-case rather than average prediction errors.

\subsection{Augmented latent state used in our implementation}
\label{app:augmented}
Our implementation's dynamics model, actor, and critic operate on the augmented latent state $x_t := (h_t,\tilde{s}_t)$, where $h_t = f_\theta(h_{t-1},\tilde{s}_{t-1},a_{t-1})$ is the dynamics model's deterministic recurrent state (Appendix~\ref{app:model_arch}) and the dynamics model predicts $\tilde{s}_t$ from $(x_{t-1},a_{t-1})$. The encoder's own GRU is internal to $e_\phi$: it is a deterministic function of $(o_{\le t},a_{<t})$ and is not part of the latent state. We show that the results of Section~\ref{sec:theory} hold with $\tilde{s}_t$ replaced by $x_t$, with the encoder-induced relation defined by $H \sim H'$ iff $x_t$ has the same distribution given $H$ as given $H'$.

\paragraph{Structure.} The augmented state is generated recursively, $x_t = g_\theta(x_{t-1},a_{t-1},\tilde{s}_t)$ with $g_\theta(x,a,s) = (f_\theta(x,a),s)$, where $\tilde{s}_t\sim e_\phi(\cdot\mid H_t)$ uses fresh noise. Since $h_t$ is a deterministic function of $(x_{t-1},a_{t-1})$, on which both the dynamics model and the adversary condition (the latter through $x_{<t}$), both place the same point mass on $h_t$; the density ratio in \eqref{eq:model_loss} therefore reduces to the ratio of densities of $\tilde{s}_t$, which is what we compute. The reward model conditions on $\tilde{s}_t$, a function of $x_t$, which is a restricted model class.

\paragraph{Lemma~\ref{lem:sufficiency}.} Let $H^i\sim H^j$, and fix an action $a$ and a value $x$ in the common support of $x_t$. Given $(H,x_t=x,a)$, the next state is $(f_\theta(x,a),\tilde{s}')$ with $\tilde{s}'\sim Q_\phi(\cdot\mid H,a)$ as in \eqref{eq:next-latent-mixture}, because the new encoder noise and the next observation do not depend on past latents. Dynamics sufficiency gives the same kernel $\kappa(\cdot\mid x,a)$ for both histories, so $Q_\phi(\cdot\mid H^i,a) = Q_\phi(\cdot\mid H^j,a)$, and identifiability of the per-step encoder implies that $P(\cdot\mid H^i,a)$ and $P(\cdot\mid H^j,a)$ induce the same distribution over next per-step encoder distributions $e_\phi(\cdot\mid H')$. The class of a next history $H' = (H,a,o')$ is determined by the law of $x_{t+1}$ given $H'$, which is the image of $\mathrm{Law}(x_t\mid H)\otimes e_\phi(\cdot\mid H')$ under $(x,s)\mapsto(f_\theta(x,a),s)$. Since $\mathrm{Law}(x_t\mid H^i) = \mathrm{Law}(x_t\mid H^j)$, the next-class distributions coincide. The reward condition follows as in Appendix~\ref{app:lemma_proof}.

\paragraph{Theorems~\ref{thm:exact} and~\ref{thm:approx}.} The proofs in Appendix~\ref{app:theorem_proof} go through verbatim with $Z^s_t = (x_{t-1},a_{t-1})$, which is a function of $P^s_t = (x_{<t},o_{<t},a_{<t})$, and with $r_t\perp\!\!\!\perp x_t\mid P^r_t$, which holds because $x_t$ is generated from the history and encoder noise that the environment does not observe. Proposition~\ref{prop:fixed-sigma} also applies, because the implemented reward model conditions on $\tilde{s}_t$, whose distribution has full support. Finally, $F$ in Appendix~\ref{app:envelope} may depend on $\theta$ through $h_t$ if the adversary network receives $h_t$ as input, which is why Proposition~\ref{prop:envelope} is stated for $\omega=(\theta,\phi)$ jointly.

\section{Implementation Details}
\label{app:implementation}
\subsection{Model Architecture and Policy Learning}
\label{app:model_arch}
Our model consists of four main components: an observation encoder, a latent dynamics model, a reward model, and an adversary network.

\paragraph{Observation Encoder.}
The encoder maps the agent's history up to time $t$, $(o_{\le t}, a_{<t})$, into a distribution over the latent state $\tilde{s}_t$. 
The architecture is composed of a convolutional neural network (CNN) followed by a GRU. 
Unlike Dreamer, which shares a GRU between the encoder and dynamics model, we use a dedicated GRU for the encoder.
The encoder outputs a mean vector $\mu_t$, normalized to have $\ell_2$ norm equal to one. 
We then sample $\tilde{s}_t$ from a Gaussian distribution with mean $\mu_t$ and fixed standard deviation $0.01$.
This normalization and fixing of the standard deviation introduces an information bottleneck: only a fixed amount of information can flow through the latent state, encouraging abstraction.

\paragraph{Latent Dynamics Model.}
The dynamics model resembles that of Dreamer, with both stochastic and deterministic components.
It maintains a deterministic hidden state $h_t$ via a GRU and outputs the parameters of a Gaussian distribution over $\tilde{s}_t$.
The mean is again normalized to unit norm, while the variance is learned but constrained to be isotropic (identical across latent dimensions). 

\paragraph{Reward Model.}
Our reward model differs slightly from Dreamer’s: it conditions only on the stochastic latent state $\tilde{s}_t$ (and the current action $a_t$), not on the deterministic hidden state $h_t$.
It outputs the parameters of a Gaussian distribution over rewards $r_t$.

\paragraph{Adversary Network.}
Finally, the adversary network shares a similar architecture with the encoder (CNN + GRU). The adversary additionally conditions on \(s_t\) through an auxiliary input before the GRU, and predicts the \emph{next} latent state $\tilde{s}_{t+1}$ instead of the current one.  Similar to the dynamics model, the adversary network outputs a unit-norm mean and an isotropic variance vector. During the model update, the adversary's parameters are frozen, but gradients propagate through its inputs and targets to the encoder (Appendix~\ref{app:envelope}).

\paragraph{Actor.} The actor is an MLP that conditions on the concatenation of the stochastic latent $\tilde{s}_t$ and the dynamics model's deterministic state $h_t$, and outputs a tanh-transformed Gaussian over actions. As in DreamerV1, it is trained by backpropagating $\lambda$-returns through imagined rollouts of the learned dynamics ($H = 15$), started from every latent state in the sampled batch. Unlike DreamerV1, the actor loss includes an entropy bonus with fixed coefficient 1, as in DreamerV2.

\paragraph{Critic.} The critic is an MLP that takes the same input as the actor and outputs a scalar value estimate. It is trained by squared-error regression onto $\lambda$-returns ($\lambda = 0.95$, $\gamma = 0.99$) computed from the reward model's mean prediction. Unlike DreamerV1, bootstrap values come from a target critic, as in DreamerV2, which is updated by a hard copy every 50 updates. Actor and critic gradients are never applied to world-model parameters.

\subsection{Model Training Algorithm}
\label{app:model_training}
\begin{algorithm}[H]

\caption{VIBES Model Training Algorithm}
\label{alg:model_training}
\DontPrintSemicolon
\SetKwInput{KwInputs}{Inputs}
\SetKwInput{KwModels}{Models}
\SetKwInput{KwHyper}{Hyperparams}
\SetKwInput{KwInit}{Initialize}

\KwInputs{Replay buffer $\mathcal{D}$ with sequences $(o_{0:T}, a_{0:T-1}, r_{0:T-1})$}
\KwModels{Encoder $e_\phi$: $(o_{\le t}, a_{<t}) \mapsto \tilde{s}_t \sim \mathcal{N}(\mu_t, \sigma_e^2 I)$ with $\|\mu_t\|_2 = 1$ and fixed $\sigma_e^2$; \\
\hspace{1.6em} Deterministic state update $h_t = f_\theta(h_{t-1}, \tilde{s}_{t-1}, a_{t-1})$ (GRU); \\
\hspace{1.6em} Dynamics $p_\theta(\tilde{s}_t \mid h_{t-1}, \tilde{s}_{t-1}, a_{t-1})$: Gaussian with unit-norm mean, isotropic variance; \\
\hspace{1.6em} Reward $p_\theta(r_t \mid \tilde{s}_t, a_t)$: Gaussian; \\
\hspace{1.6em} Adversary $\hat{p}_\psi(\tilde{s}_t \mid \tilde{s}_{<t}, o_{<t}, a_{<t})$: Gaussian with unit-norm mean, isotropic variance}
\KwHyper{Learning rates $\eta_\theta, \eta_\phi, \eta_\psi$ with $\eta_\psi > \eta_\phi$; batch size $B$; sequence length $T$}
\KwInit{Randomly initialize $\phi,\theta,\psi$; set $h_0 = \mathbf{0}$ }

\For{training steps $= 1,2,\dots$}{
  Sample $B$ trajectory chunks of length $T$ from $\mathcal{D}$\;

  \tcp{Encode latents with reparameterization}
  \For{$t=0$ \KwTo $T$}{
    $(\mu_t, \sigma_e^2) \gets e_\phi(o_{\le t}, a_{<t})$, enforce $\|\mu_t\|_2 \leftarrow 1$\;
    Sample $\epsilon_t \sim \mathcal{N}(0, I)$,\quad $\tilde{s}_t \leftarrow \mu_t + \sigma_e \odot \epsilon_t$\;
  }

  \tcp{Update deterministic hidden state, then compute latent dynamics and reward}
  \For{$t=1$ \KwTo $T$}{
    $h_t \leftarrow f_\theta(h_{t-1}, \tilde{s}_{t-1}, a_{t-1})$ \tcp*{GRU update}
    Compute $p_\theta(\tilde{s}_t \mid h_{t-1}, \tilde{s}_{t-1}, a_{t-1})$ \tcp*{latent transition}
    Compute $p_\theta(r_t \mid \tilde{s}_t, a_t)$ \tcp*{reward model}
  }

  \tcp{Adversary predictions (condition on full history)}
  Compute $\hat{p}_\psi(\tilde{s}_t \mid \tilde{s}_{<t}, o_{<t}, a_{<t})$ for $t=1{:}T$\;

  \tcp{Per-dimension non-positivity clipping of the ratio term}
  \For{$t=1$ \KwTo $T$}{
    \For{each latent dim $i$}{
      $\ell^{(i)}_t \leftarrow  \sum_{b=1}^B \log \dfrac{p_\theta(\tilde{s}^{(i)}_{t,b} \mid h_{t-1,b}, \tilde{s}_{t-1,b}, a_{t-1,b})}{\hat{p}_\psi(\tilde{s}^{(i)}_{t,b} \mid \tilde{s}_{<t,b}, o_{<t,b}, a_{<t,b})}$\;
      $\tilde{\ell}^{(i)}_t \leftarrow \min(\ell^{(i)}_t,\, 0)$ \tcp*{clip so term is non-positive}
    }
  }

  \tcp{Objective (minibatch average)}
  $\displaystyle
  \mathcal{L} \leftarrow \frac{1}{TB}  \sum_{t=1}^T \bigg[
    \sum_i \tilde{\ell}^{(i)}_{t}
    + \sum_{b=1}^B \log p_\theta(r_{t,b} \mid \tilde{s}_{t,b}, a_{t,b})
  \bigg]
  $\;

 \tcp{ Outer ascent ($\psi$ frozen; gradients flow through $\tilde{s}$ into $\hat{p}_\psi$; Prop.~\ref{prop:envelope}) }
 $\theta \leftarrow \theta + \eta_\theta \nabla_\theta \mathcal{L}$;\quad
  $\phi \leftarrow \phi + \eta_\phi \nabla_\phi \mathcal{L}$\;

  \tcp{Inner ascent update on adversary (maximize \eqref{eq:psi_objective})}
  $\displaystyle
  \mathcal{J}_\psi \leftarrow \frac{1}{TB}\sum_{b=1}^B \sum_{t=1}^T
  \log \hat{p}_\psi(\tilde{s}_{t,b} \mid \tilde{s}_{<t,b}, o_{<t,b}, a_{<t,b})
  $\;
  $\psi \leftarrow \psi + \eta_\psi \nabla_\psi \mathcal{J}_\psi$\;

}
\end{algorithm}

\subsection{Hyperparameters}
\label{app:hyperparams}
Table~\ref{tab:hyperparams} lists the hyperparameters used by VIBES.

\begin{table}[H]
\centering
{\fontsize{9}{11}\selectfont
\begin{tabular}{ll}
\hline
\textbf{Name} & \textbf{Value} \\
\hline
\multicolumn{2}{l}{\textit{Encoder}} \\
Latent dimension $s_{\dim}$ & 50 \\
Encoder standard deviation $\sigma_e$ & 0.01 (fixed) \\
Mean normalization & unit $\ell_2$ norm \\
Encoder GRU layers & 1 \\
Encoder layer norm & yes \\
Encoder learning rate & $10^{-4}$ \\
\hline
\multicolumn{2}{l}{\textit{Dynamics model}} \\
Hidden dimension & 200 \\
GRU layers & 1 \\
Variance & isotropic (shared across dimensions) \\
Model learning rate & $5\times10^{-4}$ \\
Dynamics model parameters & 0.30M \\
\hline
\multicolumn{2}{l}{\textit{Adversary}} \\
GRU layers & 1 \\
Variance & isotropic, per-dimension (not shared) \\
Adversary updates per model update & 1 \\
Adversary learning rate & $3\times10^{-4}$ \\
Adversary parameters & 1.87M \\
\hline
\multicolumn{2}{l}{\textit{Reward model}} \\
Reward normalization & yes \\
Minimum reward std & 0.1 \\
\hline
\multicolumn{2}{l}{\textit{Loss coefficients}} \\
Adversarial loss coefficient & 0.01 \\
Dynamics loss coefficient & 0.01 \\
Reward loss coefficient & 1 \\
\hline
\multicolumn{2}{l}{\textit{Actor and critic}} \\
Actor hidden dimension & 200 \\
Actor layers & 4 \\
Actor learning rate & $10^{-4}$ \\
Entropy coefficient $\alpha$ & 1 (fixed) \\
Imagination horizon $H$ & 15 \\
Critic hidden dimension & 200 \\
Critic layers & 3 \\
Critic learning rate & $10^{-4}$ \\
Discount $\gamma$ & 0.99 \\
Return lambda $\lambda$ & 0.95 \\
Target critic update interval & 50 (hard copy) \\
\hline
\multicolumn{2}{l}{\textit{Training}} \\
Sequence length $L$ & 50 \\
Batch size & 10 \\
Gradient steps per iteration & 100 \\
Prefill episodes & 5 \\
Image size & $84\times84$ \\
Environment steps & 500K \\
Model reset schedule & Episodes 250, 500, 750 \\
\hline
\end{tabular}                                                                                                                                   
}
\caption{VIBES hyperparameters}
\label{tab:hyperparams}
\end{table}

\subsection{DCS Configuration}
\label{app:dcs_config}
We evaluate on the Distracting Control Suite \citep{dcs} at the easy difficulty tier with dynamic distractors, and enable all three distractor types (background, camera, and color) simultaneously. Background distractors are drawn from the DAVIS dataset: following the package's easy-tier default, four videos are drawn from the DAVIS training split for training and four from the validation split for evaluation, so the evaluation distractors are held out and disjoint from those seen during training. Observations are $84 \times 84$ RGB images. Action repeat is 2 for Walker-walk and Finger-spin, 4 for Cheetah-run, Reacher-easy, and Ball-in-cup-catch, and 8 for Cartpole-swingup, giving episode lengths of $1000/\text{action repeat}$ steps. All methods are trained for 500K environment steps.

\subsection{Realistic ManiSkill Configuration}
\label{app:maniskill-protocol}

We use the released HRSSM Realistic ManiSkill environment and its pool of 88 Matterport3D \citep{matterport3d} indoor scenes. Agents receive $64\times64$ RGB observations and use the seven-dimensional \texttt{pd\_ee\_delta\_pose} controller. We use the released dense rewards and an action repeat of one. Episodes are truncated after 100 actions, and VIBES does not learn a continuation model. Lift Cube uses the released reward and fixed initial state. Turn Faucet uses faucet model 5021. Training and evaluation sample the same scene pool.

 VIBES, No Adv. Obj., and DreamerV3 use six training seeds and ten stochastic evaluation episodes per seed.  All three receive about 100K learner updates, corresponding to 100K environment interactions for VIBES and No Adv. Obj. and 203K for DreamerV3. Parameter counts exclude target and EMA copies.

The rows marked with $\dagger$ in Table~\ref{tab:realistic-maniskill} are taken from HRSSM Table~2: TIA, Denoised MDP, RePo, and HRSSM. We ran the remaining experiments. The DreamerV3-1M and DreamerV3-12M rows use the official DreamerV3 implementation, and the VIBES and No Adv. Obj. rows use our implementation.

\subsection{Swimmer Configuration}
\label{app:swimmer}
We build on the \emph{Swimmer} environment from the DeepMind Control Suite \citep{dmc}, a procedurally generated $k$-link planar snake robot moving in a viscous fluid, and set $k = 100$. The physical state is 204-dimensional: the 3 position coordinates of the head and the 99 joint angles, plus their velocities. The agent receives a 401-dimensional vector observation, consisting of the 99 joint angles, the 2-dimensional nose-to-target vector, and the 300-dimensional body velocities; no images are rendered. The target location is re-sampled each episode, and the reward decays smoothly with the nose-to-target distance according to a Lorentzian function. We increase the actuator gear 100$\times$ above the stock \texttt{dm\_control} swimmer (from $5\times10^{-4}$ to $5\times10^{-2}$), since the default torque scale produces poor, ineffective locomotion gaits at 100 links.

Directly controlling a 99-dimensional action space is impractical and not the focus of this study. Instead, the agent outputs the four parameters $[A,\,c_1,\,c_2,\,c_3]$ of a spatiotemporal sinusoid, and the torque applied to joint $i$ at timestep $t$ is $\tau_i(t) = A \cdot \sin(c_1 i + c_2 t) + c_3$. Each parameter is scaled for stable control. We use an action repeat of 10 to prevent the agent from changing the sinusoid parameters at unrealistically high frequencies.

VIBES, Recon.\ Obj., and DreamerV3 are each trained for 500K environment steps with 4 seeds.

\subsection{Diagnostics of the Adversary Approximation}
\label{app:adversary-gap}

\begin{figure}[H]
\centering
\includegraphics[width=\linewidth]{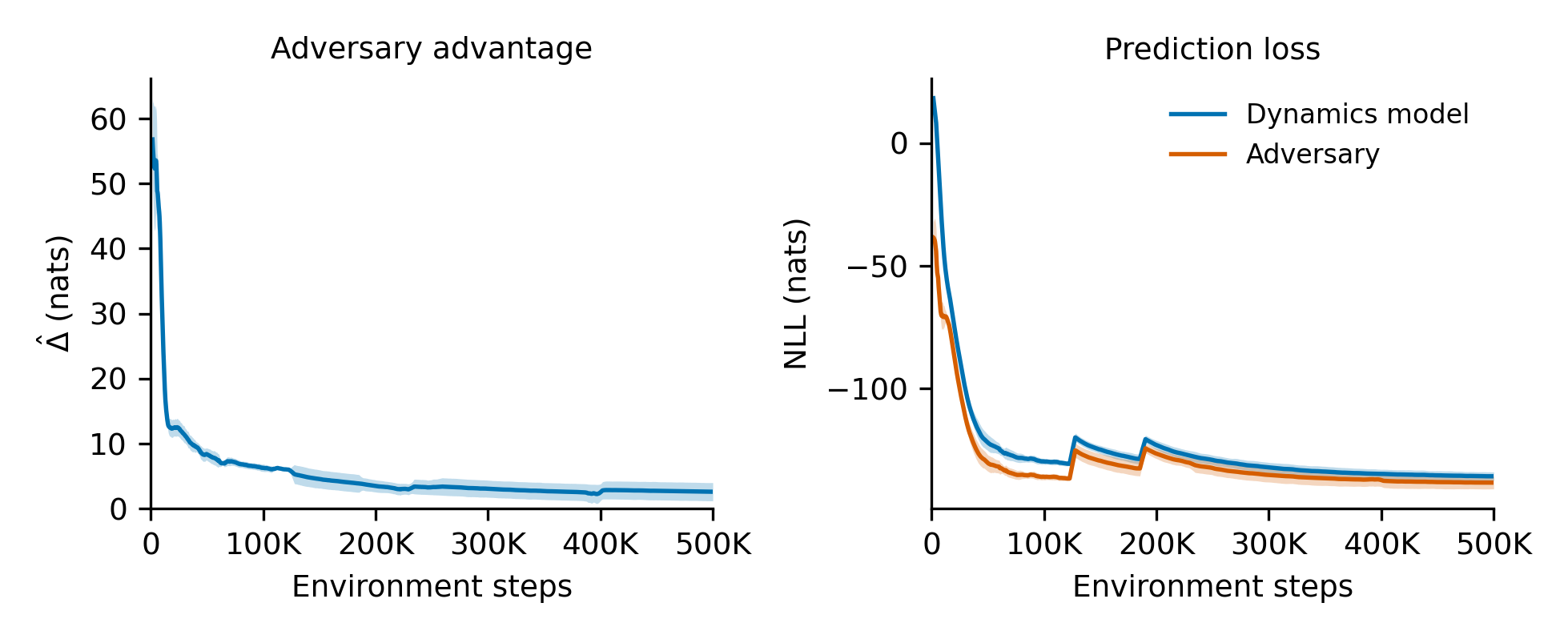}
\caption{\textbf{Adversary--dynamics likelihood gap on Cheetah-run.} (Left) $\hat{\Delta}$ = dynamics NLL $-$ adversary NLL; positive values mean the history-conditioned adversary predicts the next latent better than the dynamics model. (Right) The two NLLs separately; spikes at 125K and 187.5K steps are scheduled model resets (Table~\ref{tab:hyperparams}). NLLs are summed over latent dimensions and averaged over batch and time. Lines show the mean over 5 seeds and shading the 95\% CI; curves are smoothed with a 20-episode rolling mean.}
\label{fig:adversary-gap}
\end{figure}

To assess the bilevel approximation (Section~\ref{sec:model-objective}), we track the adversary--dynamics likelihood gap $\hat{\Delta}$ = dynamics NLL $-$ adversary NLL over training on Cheetah-run (Fig.~\ref{fig:adversary-gap}). The gap shrinks by roughly 15$\times$ over training, with its mean and 95\% CI positive throughout, so the adversarial term supplies gradient for the entire run. Both losses decrease (Fig.~\ref{fig:adversary-gap}, right), so the narrowing reflects the dynamics model catching up to the adversary rather than the adversary degrading. The gap remains positive at the end of training, consistent with Proposition~\ref{prop:fixed-sigma}.

\subsection{DCS Learning Curves}
\label{app:dcs-learning-curves}

\begin{figure}[H]
\centering
\includegraphics[width=\textwidth]{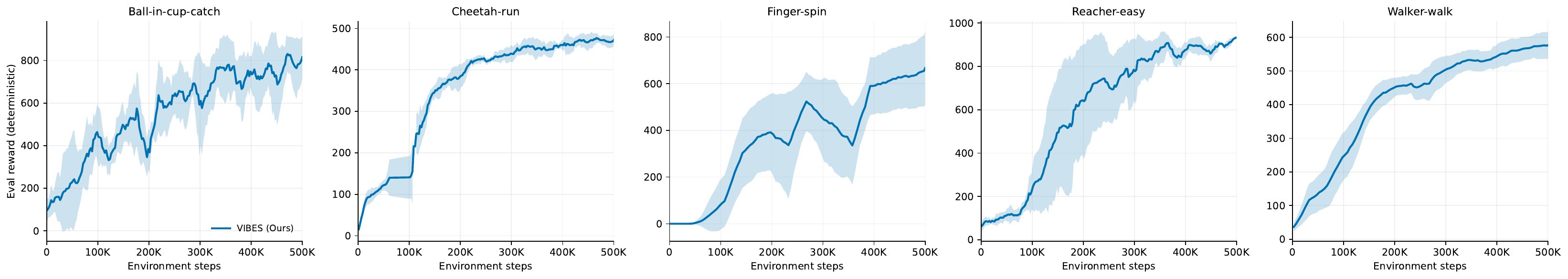}
\caption{VIBES learning curves on five Distracting Control Suite tasks. Mean evaluation return over 5 seeds versus environment steps, with 95\% confidence intervals shaded.}
\label{fig:dcs-learning-curves}
\end{figure}

\section{Statistical Significance Tests}
\label{app:significance-tests}

We compare VIBES with every other method on each task using Welch's two-sample $t$-test, which does not assume the two methods have equal variance. Seed counts are 5 for our Distracting Control Suite runs and 6 for our ManiSkill runs. Results taken from prior work use 10 seeds (RSAC, CURL, CoRe), 5 (SAC+RAD, QT-Opt+RAD, QT-Opt+DrQ), 6 (RePo, HRSSM) and 3 (TIA, Denoised MDP). Degrees of freedom follow the Welch--Satterthwaite approximation. We report two-sided $p$-values without correction for multiple comparisons; the mean-return comparisons on DCS remain significant under Holm correction. A $p$-value above 0.05 means the difference cannot be resolved with the available seeds; it does not establish that the methods are equivalent.

\begin{table}[H]
\centering
{\fontsize{7.6}{8.4}\selectfont
\renewcommand{\arraystretch}{0.9}
\begin{tabular}{lccccccc}
\hline
VIBES vs. & Mean & BiC-Catch & C-swingup & C-run & F-spin & R-easy & W-walk \\
\hline
SAC+RAD$\dagger$    & \textbf{<.001} & \textbf{<.001} & \textbf{<.001} & \textbf{0.001} & \textbf{0.015} & \textbf{<.001} & 0.078 \\
QT-Opt+RAD$\dagger$ & \textbf{<.001} & \textbf{0.003} & \textbf{<.001} & \textbf{<.001} & 0.057 & \textbf{0.002} & \textbf{<.001} \\
QT-Opt+DrQ$\dagger$ & \textbf{<.001} & \textbf{<.001} & \textbf{<.001} & \textbf{<.001} & \textbf{0.003} & \textbf{<.001} & \textbf{<.001} \\
RSAC$\dagger$       & \textbf{<.001} & \textbf{<.001} & \textbf{<.001} & \textbf{<.001} & \textbf{<.001} & \textbf{<.001} & 0.030 \\
CURL$\dagger$       & \textbf{<.001} & \textbf{<.001} & \textbf{<.001} & \textbf{<.001} & 0.818 & \textbf{<.001} & \textbf{0.040 ($-$)} \\
CoRe$\dagger$       & \textbf{<.001} & 0.446 & \textbf{<.001} & \textbf{0.020} & 0.722 ($-$) & \textbf{<.001} & \textbf{0.004 ($-$)} \\
SAR                 & \textbf{<.001} & \textbf{<.001} & \textbf{<.001} & \textbf{0.001} & \textbf{0.001} & \textbf{<.001} & \textbf{<.001} \\
DBC                 & \textbf{<.001} & \textbf{<.001} & \textbf{<.001} & \textbf{<.001} & \textbf{0.001} & \textbf{<.001} & \textbf{<.001} \\
HRSSM               & \textbf{0.022} & 0.118 & 0.154 & 0.135 & 0.901 ($-$) & \textbf{<.001} & 0.489 ($-$) \\
DreamerV3-12M       & \textbf{<.001} & \textbf{<.001} & \textbf{<.001} & \textbf{<.001} & \textbf{0.002} & \textbf{<.001} & 0.400 \\
No Adv. Obj.        & \textbf{0.002} & \textbf{0.034} & 0.239 & \textbf{0.018} & 0.077 & 0.699 & 0.584 \\
Recon. Obj.         & \textbf{<.001} & \textbf{<.001} & \textbf{0.011} & 0.163 & \textbf{0.027} & \textbf{0.016} & 0.311 ($-$) \\
\hline
\end{tabular}
}
\caption{Welch's $t$-test $p$-values, VIBES vs. each method on the Distracting Control Suite. Bold: significant difference, VIBES has the higher mean. Bold ($-$): significant difference, VIBES has the lower mean. Plain: no significant difference ($p \geq 0.05$), with ($-$) marking where VIBES has the lower mean.}
\label{tab:dcs-pvalues}
\end{table}

\begin{table}[H]
\centering
\begin{tabular}{lcc}
\hline
VIBES vs. & Lift Cube & Turn Faucet \\
\hline
TIA$\dagger$          & 0.089 & \textbf{<.001} \\
Denoised MDP$\dagger$ & \textbf{<.001} & \textbf{<.001} \\
RePo$\dagger$         & \textbf{<.001} & \textbf{<.001} \\
HRSSM$\dagger$        & \textbf{<.001} & \textbf{<.001} \\
DreamerV3-1M          & \textbf{0.049} & \textbf{<.001} \\
DreamerV3-12M         & 0.254 ($-$) & 0.105 ($-$) \\
No Adv. Obj.          & 0.360 & \textbf{0.031} \\
\hline
\end{tabular}
\caption{Welch's $t$-test $p$-values, VIBES vs. each method on Realistic ManiSkill. Bold: significant difference, VIBES has the higher mean. Plain: no significant difference ($p \geq 0.05$), with ($-$) marking where VIBES has the lower mean.}
\label{tab:maniskill-pvalues}
\end{table}

\section{Related Work}
\label{app:relatedwork}

Several classes of representation-learning methods have been proposed to learn useful representations of high-dimensional observations. We review the classes most relevant to our setting and discuss the key algorithms that belong to each.

\noindent\textbf{1. Reconstruction-Based Methods.} These methods encode latents that can be decoded back into the original observation. Because the objective rewards faithful reconstruction of every pixel, these representations have no incentive to discard distractors. As a result, they perform poorly in environments with heavy distractor content. Representative algorithms are PlaNet \citep{planet}, and the Dreamer family of algorithms \citep{dreamer,dreamerv2,dreamerv3,dreamerv4}.

\noindent\textbf{2. Contrastive and Self-Supervised Methods.} These methods define an anchor observation, a positive example, and one or more negative examples, and train a lower-dimensional encoding via a contrastive loss that pulls the anchor and positive together while pushing the anchor and negatives apart. The choice of what counts as a positive determines what the representation ends up invariant to, and this choice can lead to problematic outcomes. We look at three representative examples:

CURL \citep{curl} (Contrastive Unsupervised Representations for Reinforcement Learning) treats two randomly-cropped versions of the same observation as a matching pair, and trains the encoder to recognize them as the same while telling them apart from crops of other observations.

CoRe \citep{core} (Contrastive Recurrent State-Space Model) defines its matching pairs temporally instead: the anchor is the real encoding of the current observation, and the positive is the predicted latent for that step --- computed from the past latent and past action, before the real observation is seen --- with negatives drawn from other real and predicted encodings in the same training batch. Since the positive is generated by predicting forward from the past latent and action, what survives in the latent is whatever is temporally predictable.

RAD \citep{rad} (Reinforcement Learning with Augmented Data) drops the contrastive loss, and the whole notion of a matching pair, entirely. It simply replaces the clean observation with a randomly-distorted one and trains the ordinary RL objective on that. Because the same underlying transition is repeatedly redrawn with a different random distortion applied each time, the network is pushed toward invariance to that distortion simply by having to fit one consistent output despite the varying input.

In all three cases, what stays in the latent is decided by something other than the reward: an augmentation choice, or a notion of temporal predictability, not by whether the discarded information actually mattered for the task. These methods may discard information the agent needed, or just as easily retain information that was irrelevant.

\noindent\textbf{3. Explicit Factorization.} These methods split the latent into separately labeled pieces and only hand the policy the piece that's supposed to matter. We look at two representative examples:

TIA \citep{tia} (Task Informed Abstractions) splits the latent into a task feature and a distractor feature. Both features cooperate to reconstruct the full image. The policy only sees the task feature. This split assumes the task and distractor evolve completely independently; the authors call this assumption ``too strong'' themselves, and it breaks whenever a distractor actually interacts with the task instead of sitting apart from it.

Denoised MDP \citep{denoisedmdp} generalizes this to two axes instead of one (controllable vs. uncontrollable, crossed with reward-relevant vs. not), keeping only the controllable-and-reward-relevant quadrant. It only knows how to handle a small set of pre-defined dependency patterns between signal and noise. If a distractor and the task signal share a common cause, something driving both at once, that case falls outside what the method handles, and it has no clean way to separate them.

Both methods build a fixed factorization structure ahead of time and rely on the environment actually matching it. When it does, the split works; when task and distractor interact instead of evolving independently, both degrade, with no mechanism to detect or correct for that at run time.

\noindent\textbf{4. Bisimulation Metrics.} These methods target the same signal we do --- reward and dynamics. But they get there through metric approximations whose fixed points need not correspond to bisimulation relations. We look at two representative examples:

DBC \citep{zhanglearning} (Deep Bisimulation for Control) trains an encoder so the distance between two latent codes matches a bisimulation-metric target: reward difference plus how differently the two states are predicted to evolve. The metric is defined recursively: calculating the distance between two states' predicted next-states requires already knowing the bisimulation distance between the current states, the very thing being solved for. DBC's implementation substitutes this ground distance with Euclidean distance instead, an incorrect approximation that shows up in practice: DBC shows poor performance when tested on out-of-distribution distractors \citep{tpc}.

HRSSM \citep{hrssm} (Hybrid Recurrent State-Space Model) combines a masking strategy with a bisimulation loss on top of DreamerV3. HRSSM's own analysis admits each half is individually insufficient. Masking alone can accidentally hide task-critical detail along with the distractor. The bisimulation loss alone isn't guaranteed to filter anything if the dynamics model it depends on is imperfect. The two are combined on the empirical expectation that they cover each other's gaps. RePo \citep{repo} introduced Realistic ManiSkill, and HRSSM \citep{hrssm} later evaluated its model on the same benchmark. Section~\ref{sec:realistic-maniskill} adds comparisons with the official DreamerV3 1M and 12M configurations.

Note: TIA, HRSSM, and CURL each rely on an image-specific operation --- pixel-space mask composition, spatio-temporal masking, and random crop, respectively --- so none of them can be adapted to vector-based inputs the way our method can.

 In contrast, VIBES does not approximate a bisimulation distance. Its objective decomposes exactly into the encoder's violation of reward and dynamics sufficiency plus model misfit (Theorem~\ref{thm:approx}), and whenever exact sufficiency is attainable, every global optimum induces a bisimulation relation over histories (Theorem~\ref{thm:exact}). It requires no domain-specific augmentation heuristics and no pixel reconstruction.  The objective rewards retaining everything needed to predict reward and dynamics, and the encoder's fixed-capacity bottleneck pressures it to drop the rest.

\section{Latent Space Analysis}
\label{app:latent-analysis}

\begin{sidewaysfigure}[p]
  \centering
  \includegraphics[width=\linewidth]{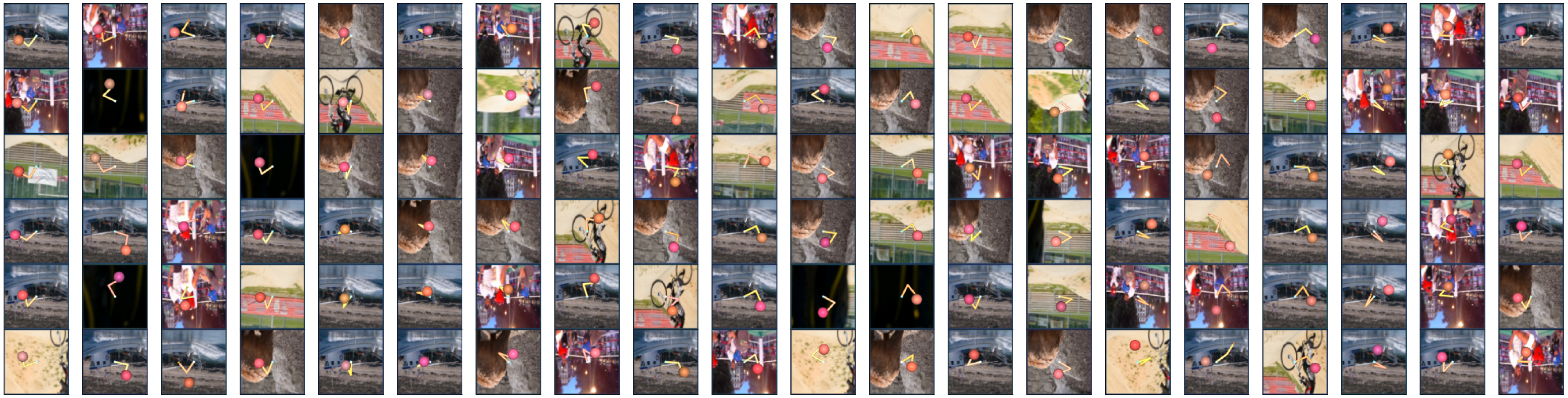}
  \captionsetup{width=.9\linewidth}
  \caption{\textbf{Nearest-neighbor analysis in latent space (all 20 queries).}
  For each of 20 query observations (left-most image in each row), we display the five nearest neighbors in the learned latent space beneath it (standardized $L_2$ distance).
  The visualization shows invariance to visual distractors and clustering of rotationally symmetric arm–goal configurations.}
  \label{fig:latent-nn-all-20}
\end{sidewaysfigure}

\begin{figure}[p]
  \centering
  \includegraphics[height=0.85\textheight]{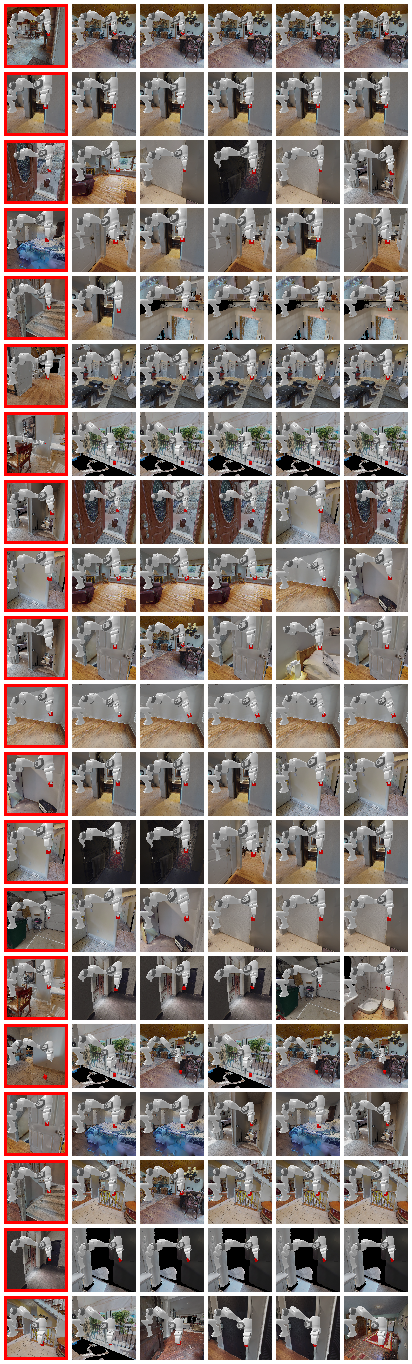}
  \captionsetup{width=.9\linewidth}
  \caption{\textbf{Nearest latent neighbors on Lift Cube for VIBES (all 20 queries).} Each row shows a query frame (outlined in red) followed by its five nearest neighbors in latent space. Neighbors match the query's gripper and cube configuration but mostly come from different scenes.}
  \label{fig:lift-cube-nn-vibes-all20}
\end{figure}

\begin{figure}[p]
  \centering
  \includegraphics[height=0.85\textheight]{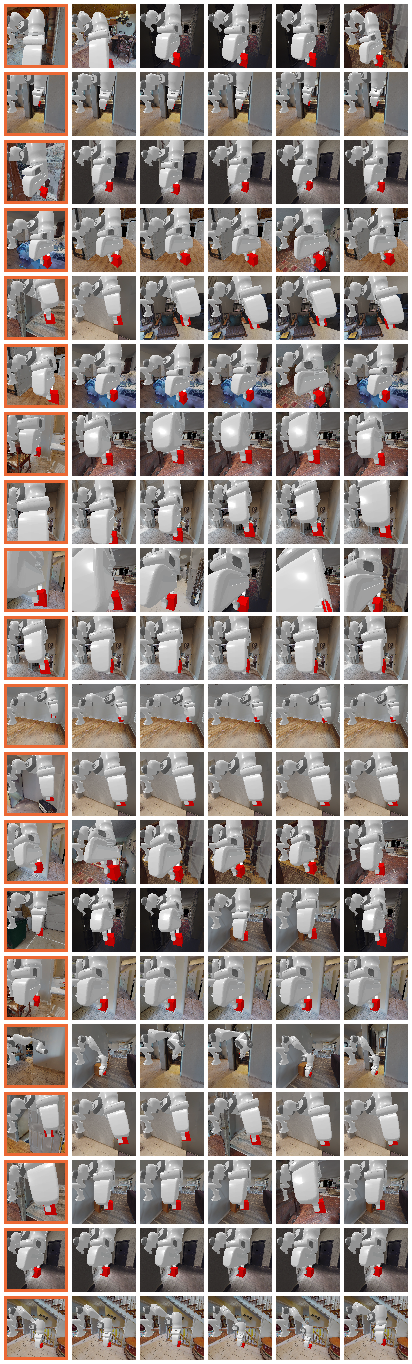}
  \captionsetup{width=.9\linewidth}
  \caption{\textbf{Nearest latent neighbors on Lift Cube for DreamerV3-12M (all 20 queries).} Layout as in Figure~\ref{fig:lift-cube-nn-vibes-all20}. Neighbors frequently share the query's scene, even when the gripper and cube configuration differ.}
  \label{fig:lift-cube-nn-dreamer-all20}
\end{figure}

This appendix shows the full nearest-neighbor results summarized in Section~\ref{sec:latent-analysis}. Figure~\ref{fig:latent-nn-all-20} shows all 20 Reacher-easy queries. Figures~\ref{fig:lift-cube-nn-vibes-all20} and~\ref{fig:lift-cube-nn-dreamer-all20} show 20 Lift Cube queries each for VIBES and DreamerV3-12M. In every figure, the leftmost image in a row is the query, and the remaining five are its nearest neighbors in latent space under standardized L2 distance, excluding frames from the query's own episode.

\section{LLM Usage}

In this work, we used generative AI coding assistants to implement competing baseline algorithms and to write the code for some of our experiments, including scripts that run them and format the resulting data. The authors read and verified the baseline implementations and checked all data presented in the paper against the experiment logs. Coding assistants also helped compute the learned-parameter counts in Table~\ref{tab:realistic-maniskill}, and the authors verified the counting methodology. The original theoretical framework, hypotheses, and experimental design, as well as the implementation of VIBES itself, were developed without generative AI assistance. Generating datasets, translation, and qualitative data analysis are not applicable to this work. Additionally, we used generative AI tools to draft and edit parts of the paper for readability and concision, write figure captions and appendix text, format LaTeX tables and references, and locate and verify details of prior work. We used a generative AI assistant (Claude) to help revise the statements and proofs (Lemma~\ref{lem:sufficiency}, Theorems~\ref{thm:exact}--\ref{thm:approx}, Propositions~\ref{prop:envelope}--\ref{prop:fixed-sigma}, and Appendices~\ref{app:lemma_proof}--\ref{app:augmented}), and to help carry out and interpret the statistical tests (Appendix~\ref{app:significance-tests}) and the training diagnostics (Appendix~\ref{app:adversary-gap}), including analysis scripts and figures. The authors checked all of these and take responsibility for them. We take responsibility for the final content of this work, including text, claims, and artifacts produced with the aid of generative AI.
\end{document}

%% file: math_commands.tex
\usepackage{amsmath,amsfonts,bm}

\def\eqref#1{equation~\ref{#1}}

\def\1{\bm{1}}

\DeclareMathAlphabet{\mathsfit}{\encodingdefault}{\sfdefault}{m}{sl}
\SetMathAlphabet{\mathsfit}{bold}{\encodingdefault}{\sfdefault}{bx}{n}

